\documentclass{article}

 \usepackage[dandb, final]{neurips_2025}
\usepackage[dvipsnames, table]{xcolor}

\usepackage{graphicx}
\usepackage{amsmath}
\usepackage{amssymb}
\usepackage{tcolorbox}
\usepackage{booktabs}
\usepackage{mathtools}
\usepackage[accsupp]{axessibility}  % Improves PDF readability for those with disabilities.
\usepackage{color}
\usepackage{array}
\usepackage{pdflscape}
\usepackage[longtable]{multirow}
\usepackage{longtable}
\usepackage{tabularray}
\usepackage{booktabs}
\usepackage{times}
\usepackage{epsfig}
\usepackage{amsmath}
\usepackage{example}
\usepackage{listings}
\usepackage{bbm}
\DeclareMathAlphabet\mathbfcal{OMS}{cmsy}{b}{n}
\usepackage{makecell}
\usepackage{siunitx}
\usepackage{svg}
\usepackage{float}
\usepackage{comment}
\usepackage{lipsum}
\usepackage{stfloats}
\usepackage{multicol}
\usepackage{multirow}
\usepackage{bm}
\usepackage{etoolbox}
\usepackage{icomma}
\usepackage{array}
\usepackage{tabulary}
\usepackage{xcolor}
\usepackage{paralist}
\usepackage{booktabs}
\usepackage{adjustbox}
\usepackage{caption}
\usepackage{subcaption}
\usepackage{arydshln}
\usepackage{rotating}
\usepackage{enumitem}

\definecolor{gray}{rgb}{0.3,0.3,0.3}
\definecolor{blue}{rgb}{0,0.5,1}
\definecolor{mask_red}{rgb}{1,0,0.8}
\definecolor{green}{rgb}{0.2,1,0.2}
\definecolor{rblue}{rgb}{0,0,1}
\definecolor{lightblue}{HTML}{6495ed}
\definecolor{lightred}{HTML}{F19C99}

\definecolor{graytablerow}{gray}{0.6}

\usepackage{pifont}% http://ctan.org/pkg/pifont
\usepackage{tabu}
\usepackage[pro]{fontawesome5}

\usepackage{tikz}

\tcbuselibrary{listingsutf8}
\usepackage[pro]{fontawesome5}

\newcommand{\ie}{\emph{i.e.}}
\newcommand{\eg}{\emph{e.g.}}

\usepackage{wrapfig}
\definecolor{cvprblue}{rgb}{0.21,0.49,0.74}
\usepackage[pagebackref,breaklinks,colorlinks]{hyperref}
\hypersetup{colorlinks, citecolor=cvprblue}%teal

\usepackage[capitalize]{cleveref}
\crefname{section}{Sec.}{Secs.}
\Crefname{section}{Section}{Sections}
\Crefname{table}{Table}{Tables}
\crefname{table}{Tab.}{Tabs.}

\title{
\Large mmWalk: Towards Multi-modal Multi-view Walking Assistance % Embodied AI
}

\author{
\textbf{Kedi Ying}$^{1}$\thanks{Equal contribution. $^{\dagger}$ Project lead. $^{\ddagger}$ Corresponding author.} \qquad \textbf{Ruiping Liu}$^{1*{\dagger}}$ \qquad \textbf{Chongyan Chen}$^{5}$ \qquad \textbf{Mingzhe Tao}$^{1}$ \qquad \textbf{Hao Shi}$^{6}$ \\ \textbf{Kailun Yang}$^{3}$ \qquad \textbf{Jiaming Zhang}$^{1,3,4{\ddagger}}$ \qquad \textbf{Rainer Stiefelhagen}$^{1,2}$ \\ $^{1}$ CV:HCI, KIT \quad $^{2}$ Center for Digital Accessibility and Assistive Technology (ACCESS@KIT) \\ $^{3}$ Hunan University \quad $^{4}$ ETH Zurich \quad $^{5}$ University of Texas at Austin  \quad $^{6}$ Zhejiang University  
}

\begin{document}
% \maketitle
% ---- For twocolumn Fig 1
% \twocolumn[{%
% \renewcommand\twocolumn[1][]{#1}%
% \maketitle
% \begin{center}
%     \centering
%     \captionsetup{type=figure}
%     \includegraphics[width=1.0\textwidth,height=5cm]{example-image-a}
%     \captionof{figure}{Test caption}
%     \label{fig:banner}
% \end{center}%
% }]
% ---- For multiple figures 
% \twocolumn[{%
\renewcommand\twocolumn[1][]{#1}%
\maketitle
\begin{center}
    \centering
    \captionsetup{type=figure}
    % \captionof{figure}{Test caption}
    % \includegraphics[width=1.0\textwidth,height=5cm]{example-image-a}
    \begin{subfigure}[t]{1.0\textwidth}
        \centering
        \includegraphics[width=\textwidth]{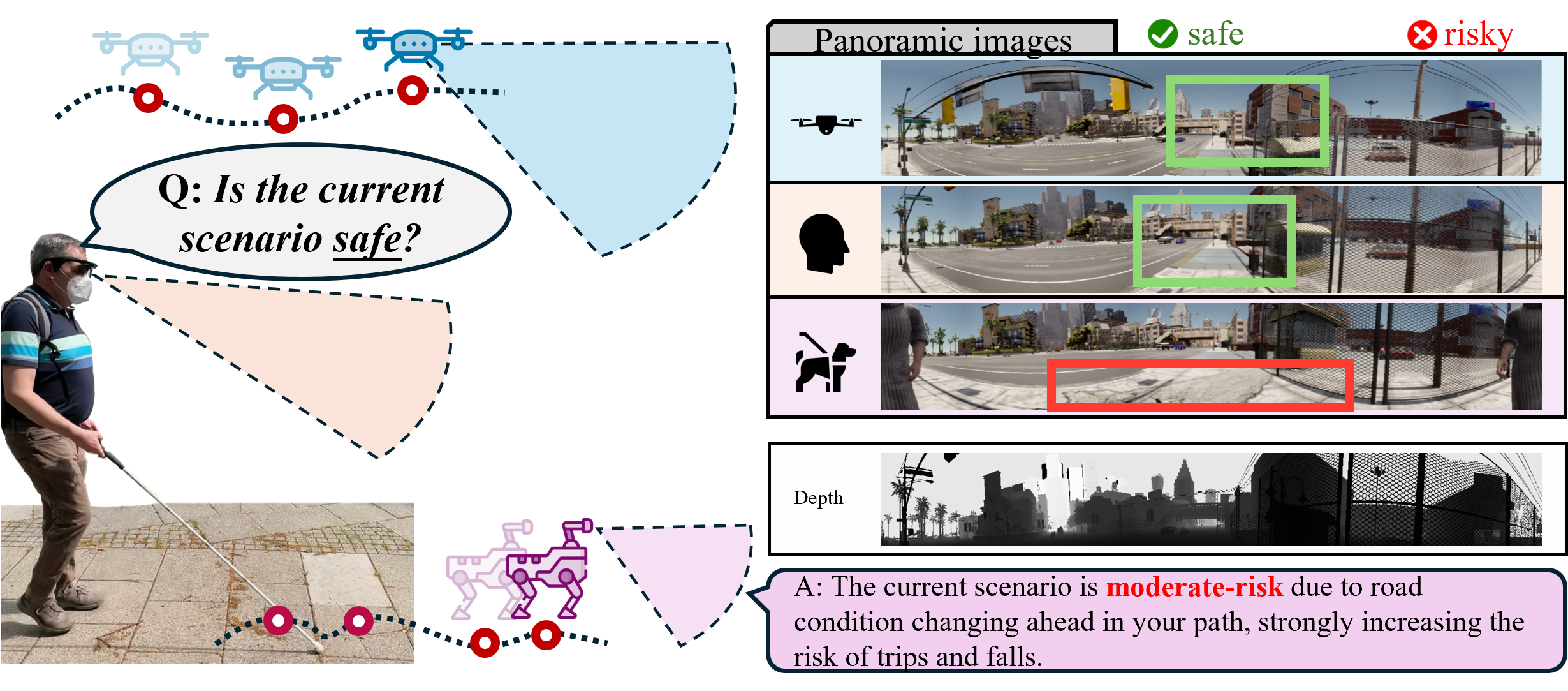}
        % \caption{TODO There is another version. figure1B.png. try one try if you want. There is another version figure1C.png. try one try if you want}
        \label{fig1-b}
    \end{subfigure}%\hfill
    \caption{\textbf{mmWalk} benchmark with multi-modal (\eg, RGB, depth), multi-view (\ie, drone, walker, guide dog), accessible (corner-case awareness with visual question answering) features.}
    \label{fig1:banner}
\end{center}%
% }]

%%%%%%%%% ABSTRACT
\begin{abstract}
% Walking assistance in field danger awareness and effective navigation for people affected by blindness and low vision (BLV) remains a challenging problem requiring a holistic multi-modal understanding of complex environments. In this work, 
Walking assistance in extreme or complex environments remains a significant challenge for people with blindness or low vision (BLV), largely due to the lack of a holistic scene understanding. Motivated by the real-world needs of the BLV community, we build \textbf{mmWalk}, %from the simulator,
a simulated multi-modal dataset that integrates multi-view sensor and accessibility-oriented features for outdoor safe navigation. Our dataset comprises $120$ manually controlled, scenario-categorized walking trajectories with $62k$ synchronized frames. 
It contains over $559k$ panoramic images across RGB, depth, and semantic modalities. 
Furthermore, to emphasize real-world relevance, each trajectory involves outdoor corner cases and accessibility-specific landmarks for BLV users. Additionally, we generate \textbf{mmWalkVQA}, a VQA benchmark with over $69k$ visual question-answer triplets across $9$ categories tailored for safe and informed walking assistance. We evaluate state-of-the-art Vision-Language Models (VLMs) using zero- and few-shot settings and found they struggle with our risk assessment and navigational tasks. We validate our mmWalk-finetuned model on real-world datasets and show the effectiveness of our dataset for advancing multi-modal walking assistance. \\ 
{\faDatabase} Data: \url{https://doi.org/10.7910/DVN/KKDXDK}. \\
{\faGithub} Code: \url{https://github.com/KediYing/mmWalk}. %\url{https://doi.org/10.7910/DVN/KKDXDK}.
\end{abstract}
\label{sec:abs}

%%%%%%%%% BODY TEXT
\section{Introduction}
\label{sec:intro}
Blindness and Low Vision (BLV) affect more than 2.2 billion people~\cite{WHO2023}, impacting their ability to travel outdoors and consequently influencing their quality of life and engagement in daily activities. One of the most critical challenges is the clichéd term of outdoor navigation. There are many outdoor navigation aids available, ranging from traditional devices to modern electronic aids to computer vision and AI assistance~\cite{naviReview1,naviReview2,navigStreetCam,navigCV}, including a significant proportion of landmark-based navigation systems~\cite{landmark01,landmark2}, and a considerable number of notable landmarks for navigating for people with BLV can be found in ATmaps statistics~\cite{ATMAPs}. Despite all of that, the survey~\cite{navSurvey} indicates that more than $63\%$ of the respondents have experienced at least one incident of injury while navigating in outdoor environments. 
Furthermore, in~\cite{obstaInjury1}, it was reported that $7\%$ of individuals with BLV experience at least one fall monthly. A model that prioritizes safety awareness is equally as important as one that ensures answer accuracy. Moreover, there are many scenarios or objects that increase the danger of the current navigation or walking path, including crossing the road, uneven ground, steps, and obstacles on the pavement~\cite{corner+cornerAssist,corner+View,corner1,blindCross}, which makes the term safety very challenging. In this context, an aid that balances hazard awareness and landmark detection is critical and more useful for the BLV community.
%Therefore, safety is as important as accuracy when it comes to outdoor navigation aids for people with BLV. 

Given these challenges, we introduce \textbf{mmWalk}, along with \textbf{mmWalkVQA}, a novel \textbf{m}ulti-view and \textbf{m}ulti-modal inclusive \textbf{Walk}ing dataset (Figure~\ref{fig1:banner}). {mmWalk} incorporates synchronized frames, referring to a single timestamp, at which multiple panoramic images (from different views) are captured, including all modalities. The frames are collected in the \textit{Carla Simulator}~\cite{carla} manually, within walker, guide dog, and drone views, capturing rich panoramic pedestrian-egocentric images including RGB, depth, semantic segmentation along with walker's action and inertial measurement unit (IMU) in $120$ trajectory path with native action among $7$ scenario categories, summing up over $559k$ images. Additionally, we defined $8$ corner cases for people with BLV among the aforementioned outdoor dangers and listed $18$ valuable navigational landmarks according to ATmaps \cite{ATMAPs}, a European standard platform for summarizing landmarks for tactile maps designed for BLV. The corner cases and landmarks, along with scenario descriptions and weather conditions, are stored in the contextual metadata. Figure~\ref{fig:overview} gives an example of a parking area scenario trajectory, with a few examples of corner cases and landmarks. In each trajectory, subsets have been further annotated through frame sampling, which were used to generate $69k$ Visual Question-Answering (VQA) for \textbf{mmWalkVQA} pairs by GPT-4o~\cite{openai2024gpt4o} in $9$ VQA-types of $3$ difficulty levels, further enabling extensive benchmarking and a series of experiments with state-of-the-art large language and vision-language models. Section~\ref{sec:dataset} describes the collection phase and the construction of the dataset, with a comprehensive presentation and deep analysis of the dataset structure and content.

\begin{figure*}[h]
    \centering
    % \includesvg[width=1\textwidth,inkscapelatex=false]
    \includegraphics[width=1\textwidth]{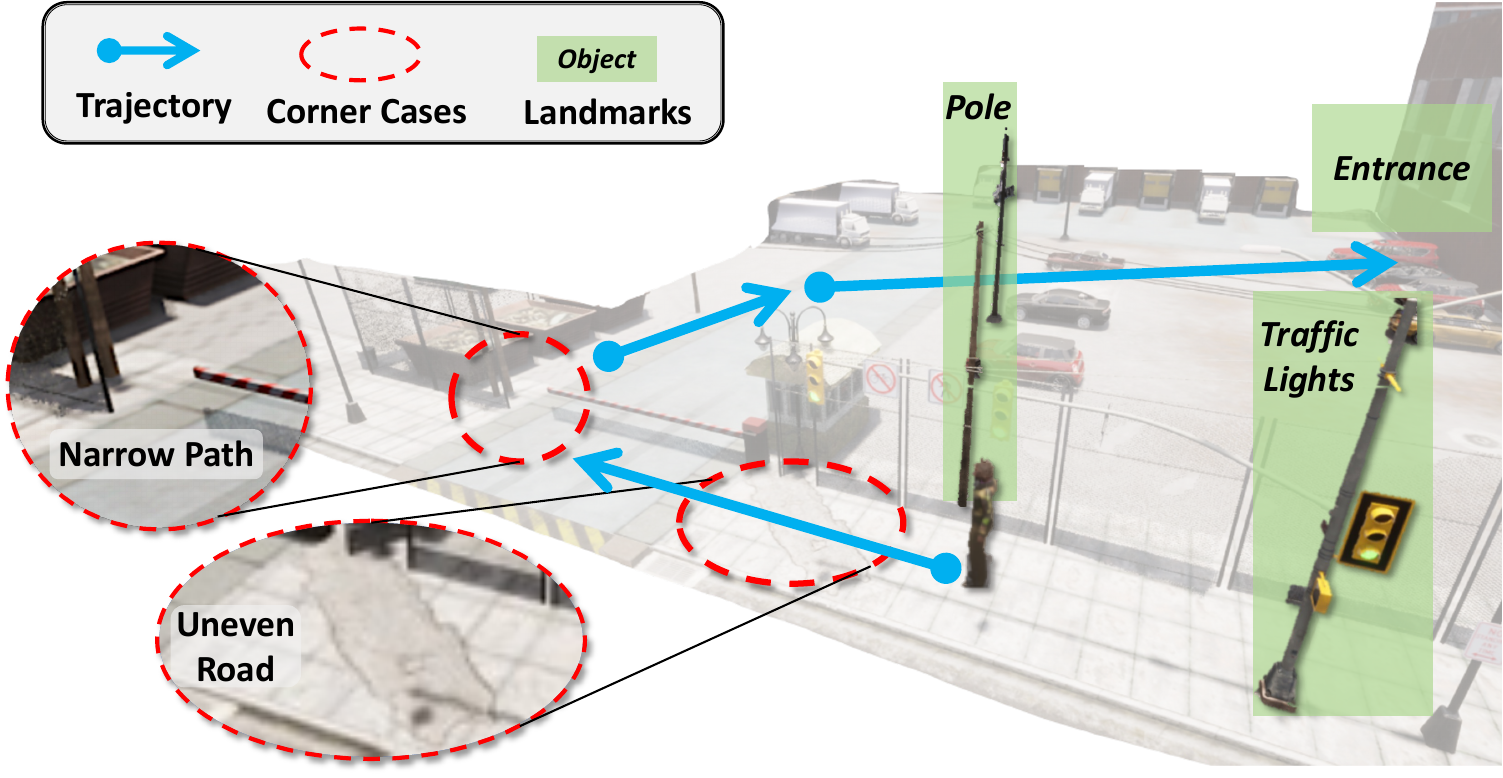}
    \caption{Visualization of a challenging walking scenario of the established mmWalk dataset. While walking in a parking area (the trajectory is the marked blue path), corner cases include \textit{narrow path} and \textit{uneven road}. The navigational landmarks for BLV include \textit{pole}, \textit{traffic light}, and \textit{entrance}.} 
    \label{fig:overview}
\end{figure*}

With mmWalk and mmWalkVQA, we provide a comprehensive analysis of existing open-source Vision-Language Models (VLMs) with multi-image capability, revealing critical limitations in the performance. With the metric of LLM-evaluation, we calculated the normalized score from GPT-4o-mini~\cite{openai2024gpt4o} upon the zero-shot and 3-shot inference results of tested models. Besides, the comparison spans multiple dimensions, including scores across different tasks (VQA types), different scenarios, and different inputs. Furthermore, our analysis demonstrates that even state-of-the-art models struggle with risk assessment and navigational tasks, which provide valuable insights for future model development and optimization, specifically targeted at assistive technologies for the BLV community. We also demonstrate the generalization of a model fine-tuned on our task when deployed for real-world outdoor visual question answering.
% and Vizwiz~\cite{gurari2018vizwiz} for real-world BLV assistance question answering tasks. %Full experimental analysis is provided in Section~\ref{sec:experiments}.
 
In summary, our main contributions are as follows:
\begin{itemize}[leftmargin=*]
\item We introduce \textbf{mmWalk}, a novel multi-view and multi-modal dataset specifically designed for inclusive walking assistance, encompassing synchronized frame data from walker, guide dog, and drone perspectives with comprehensive modalities (RGB, depth, semantic segmentation), contextual metadata, and a large size (over $559K$ images). 

\item We provide a scalable pipeline to generate BLV-oriented VQA pairs, offering an accessible and inclusive benchmark (\textbf{mmWalkVQA}) for evaluating VLMs in assistive tasks, including scene understanding, pedestrian navigation, and risk assessment for individuals with BLV.

\item We analyze the performance of many VLMs on mmWalkVQA, revealing significant limitations in their ability to reason about spatial relationships, identify hazards, and comprehend multi-view scenes from the perspective of BLV users. Cross-evaluation on the real-world dataset proves that VLMs obtain significant benefits by fine-tuning on the established mmWalk dataset.
\end{itemize}

%-------------------------------------------------------------------------
%%%%%%%%% 
\section{Related Work}
\label{related_work}
% To comprehensively contextualize our work and contribution, we present in this section an overview of prior research in assistive and navigational technologies for the BLV community, other visual aid datasets, as well as studies on special outdoor dangerous corner cases for BLV individuals.

\textbf{Walking and Navigation Assistance.}
% Navigation assistance systems for people with blindness or low vision (BLV) have evolved significantly over recent decades. 
%Traditional assistive systems focused on obstacle detection rather than navigation guidance. For example, NavBelt~\cite{nav1998} used ultrasonic sensors to detect obstacles and provide directional guidance. With advances in computer vision and artificial intelligence, camera-based navigation systems have become increasingly prominent, which can identify obstacles, understand the scene in the environment, and provide spatial awareness through audio descriptions~\cite{navigCV,navigStreetCam}
Multi-view assistance systems represent an important advancement in outdoor navigation. For example, the BLV assistant OpenMPR~\cite{multiview1} is a place recognition system utilizing multi-view image data for place matching. The Multi-view Street Scene Perception (MSSP) system~\cite{multiview2} uses multiple camera views to enhance the perception of pedestrian paths and obstacles in complex urban environments. In the navigation domain, researchers have investigated how to deploy multi-view sensors~\cite{drone1,drone2}, particularly drones, to build a multi-view navigation assistant~\cite{drone3,drone4,drone5}. Despite these technological advances, many navigation systems fail to adequately address safety-critical aspects such as identifying hazardous conditions, uneven surfaces, and temporary obstacles~\cite{navSurvey,obstaInjury1}. To overcome this limitation, the proposed mmWalk aims to cover safety-critical factors through its focus on corner cases and accessible landmarks.

\textbf{Visual Assistive Datasets.}
% There exist numerous highly regarded vision-based assistive datasets designed for the BLV community. 
Numerous visual assistive datasets have been developed for the BLV community, focusing on indoor navigation, object recognition, and text-to-speech conversion for reading assistance. Among these, VQA datasets are particularly prevalent and relevant to our work. For example, the VizWiz dataset~\cite{gurari2018vizwiz} contains over $31,000$ image-question pairs where visually impaired individuals captured images and asked questions to learn about their surroundings. The recent GuideDog dataset~\cite{kim2025guidedog} represents a significant step toward egocentric multi-modal data collection for BLV assistance, along with VQA pairs. SideGuide~\cite{sideguide} incorporates spatial and depth information with egocentric perspectives specifically for BLV users. More generally, there are pedestrian datasets that are not specifically designed for BLV users but can still be applied to BLV-related scenarios and navigation tasks, such as TBRSD~\cite{tbrsd}, X-World~\cite{xworld}, SANPO~\cite{waghmare2023sanpo}, EGO4D~\cite{ego4d}, Musohu~\cite{musohu}, SpatialLLM~\cite{ma2025spatialllmcompound3dinformeddesign}, and the space-aware instruction tuning dataset~\cite{robotassist}. Unlike existing datasets, our mmWalk dataset involves corner cases of walking scenarios that hinder the generalizability of blind assistive systems, %which focuses on the primary objective of cultivating more context-aware and responsive assistance, 
providing BLV with accessible landmarks for navigation. %, which will be detailed in Section~\ref{sec:dataset}.

\textbf{Corner Cases for BLV.}
\label{sec:corner}
% As mentioned in Section~\ref{sec:intro}, 
People who are Blind or with Low Vision (BLV) often face unique safety challenges when navigating outdoor environments. Research on BLV navigation challenges has identified several critical outdoor corner cases. \emph{Road crossing} consistently emerges as one of the most significant concerns, with studies indicating that they are among the highest-risk activities for BLV pedestrians~\cite{corner1,blindCross,cross1,cross2}. 
Additionally, \emph{uneven ground or road} outdoors is considered one of the dangerous challenges~\cite{corner1,uneven1}. Specifically, ~\cite{uneven1} specifically mentioned \emph{irregular pavements}, \emph{unknown stairways}, and \emph{roadside potholes} as dangerous factors. Moreover, a study~\cite{barrier1} discusses the challenges of dealing with \emph{barriers} on the road, including recognizing objects and getting around obstacles on the road. \cite{barrier2} points out that obstacles on the road, such as mailboxes and parked motorbikes, can make navigation much harder, creating a more \emph{narrow path} that can be walked through. Another challenge is \emph{finding entrances and exits} to houses, buildings, or underground stations~\cite{entrance1,entrance2,entrance3}. 
Maxime~\textit{et al.}~\cite{high1} have made specific reference to the dangers of \emph{obstacles in high positions} such as sagging tree branches and street vendors' awnings. 
By incorporating these corner cases into the fresh mmWalk dataset, our work aims to enable the development of more comprehensive and safety-oriented navigation assistance systems that can better address the full spectrum of challenges faced by BLV individuals in outdoor mobility.

\section{Dataset Creation}
\subsection{mmWalk Dataset}
\label{sec:dataset}

\begin{wrapfigure}[17]{r}{0.4\textwidth}
  \centering
  \vskip -2ex
  \includegraphics[width=0.4\textwidth]{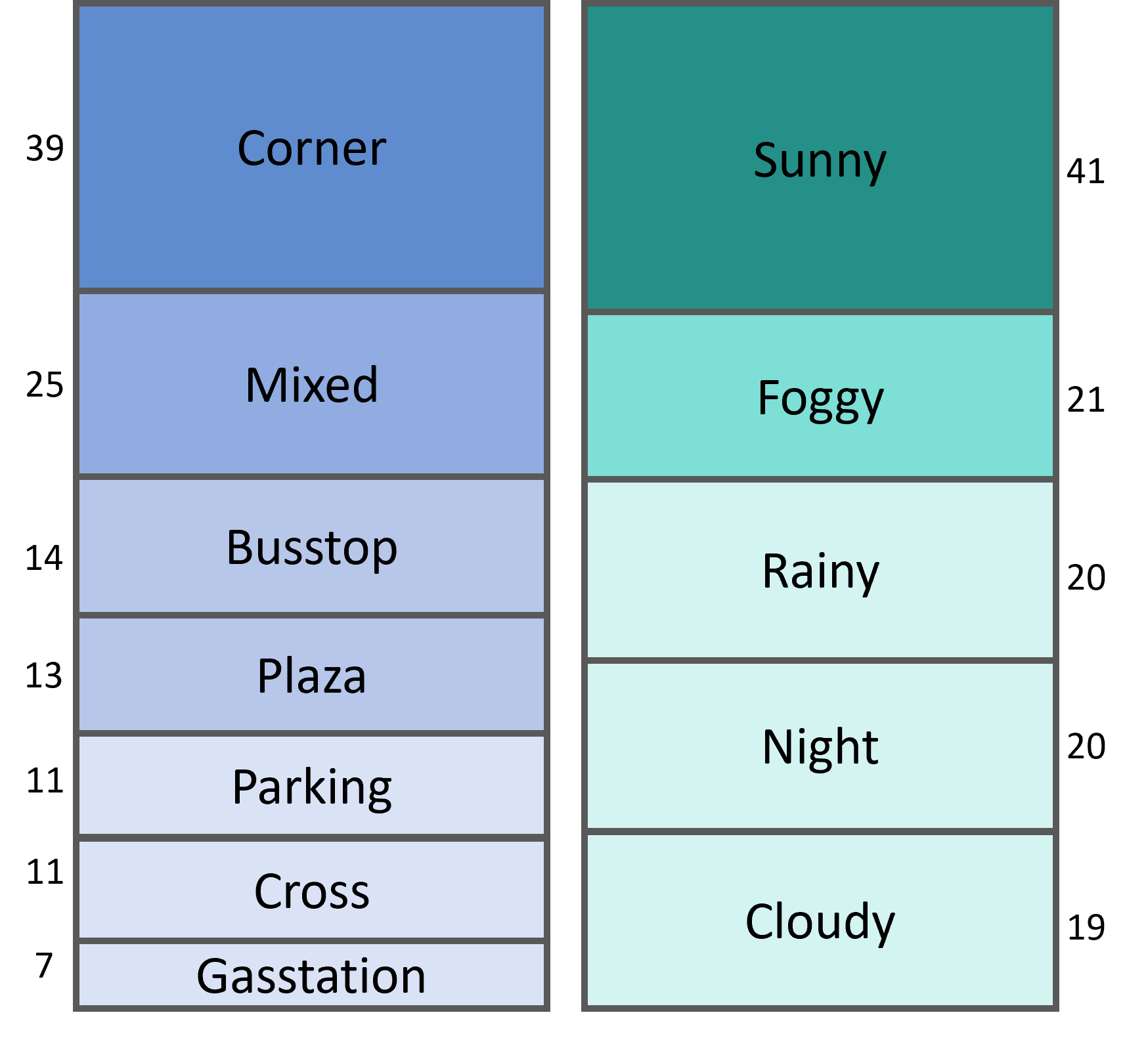}
  \caption{Trajectory numbers of each scenario (left) and weather (right).}
  \label{fig3:traj_distribute}
\end{wrapfigure}

Our \textbf{mmWalk} dataset was collected using Carla~\cite{carla}, a popular open-source simulator for autonomous driving with customization capabilities and supporting a variety of sensors, which allows us to customize the trajectory and collect an ego pedestrian dataset. mmWalk dataset contains $120$ trajectories across of $7$ scenario categories and $5$ weather conditions (Figure~\ref{fig3:traj_distribute}). The trajectories have an average of $518$ frames. There are three uniquely deployed sensor-groups for each view. With multiple views and multiple modalities including depth and semantic segmentation, mmWalk provides $62,167$ frames and over $2.5M$ single images in total, which are collated as a cubemap per frame and converted into an equirectangular panoramic image by the py360convert toolkit~\cite{py360}, summing up $559,503$ panoramic images. In addition, we labeled and stored the trajectory metadata which contains trajectory descriptions, occurred BLV corner cases, special landmarks, and the action of the ego pedestrian. More specifics of the dataset collection can be found in Appendix \ref{app1}.

%Additionally, mmWalk support the evaluation of finetuned models. % optimized for the VQA task. 
%Also, mmWalk is the only dataset that features multi-view, panorama, sequential trajectory, and egocentric image, which increase the accuracy of the dataset's information and the breadth of its application, as well as the types of VQA specifically designed and adapted for the BLV population, and a finetuned model.

\begin{wraptable}[13]{r}{0.6\textwidth}
\centering
\caption{Categories and descriptions of the corner cases.}
\label{tab:corner}
\renewcommand\arraystretch{1.2}
\setlength{\tabcolsep}{3pt}
\resizebox{0.6\textwidth}{!}{%
\begin{tabular}{llc}
\hline
\textbf{Corner Case}     & \textbf{Description}                                              & \textbf{\#Traj.} \\ \hline
Cross road in danger & No traffic light or zebra cross & 16                    \\
Cross road    & Cross the road generally              & 17 \\
Uneven road       & Ground condition changes               & 40 \\
Barrier           & Obstacles in the path blocking the way & 27 \\
Narrow path       & Walking through a narrow path          & 36 \\
Entrance locating & Find path onto a small entrance   & 22 \\
High obstacles    & High position obstacles                & 11 \\
Deadend           & Walking into a deadend                 & 5  \\ \hline
\end{tabular}%
}
\end{wraptable}

\textbf{Scenario and Weather.} mmWalk trajectories are strategically distributed across $7$ urban scenario categories and $5$ weather conditions. Figure~\ref{fig3:traj_distribute} shows the number of trajectories in each scenario and weather. The scenario represents a specific type of environment or location characterizing the overall trajectory. In the cases of the ‘Corner’ scenario, the character has no other special behavioral logic or path goal, but mainly focuses on the corner case, for example, \textit{`bypassing the obstacle on the path'}.

\textbf{Corner Cases and Landmarks.} 
As introduced in Section~\ref{sec:corner}, we identify and summarize $8$ corner cases critical for BLV users, based on prior literature, and incorporate them into mmWalk. Table~\ref{tab:corner} provides a succinct and clear description, as well as the number of trajectories containing each corner case category. Note that in the non-corner scenario, depending on our pathway design, one or even more types of corner cases will likewise appear. We can see that critical corner cases such as \textit{uneven road}, \textit{road crossing} (sum up the general crossings and dangerous situational crossings), and \textit{narrow paths} appear very frequently in the dataset. Walking into a \textit{dead-end} road occurs the least, and usually occurs together with \textit{entrance locating}. 

For accessibility landmarks, we refer to the list of the most important landmarks identified by blind people, as surveyed in ATmaps~\cite{ATMAPs}. From the list, we then select the $18$ items that appear more frequently in the simulator. The specific details of these landmarks are provided in Appendix \ref{app_annotation}.

\subsection{VQA Types and Generation}

\begin{wraptable}[14]{r}{0.5\textwidth}
\centering
\caption{Statistics of VQA types grouped by difficulty levels and categories.}
\label{tab:difficulty_levels}
\renewcommand\arraystretch{1.1}
\setlength{\tabcolsep}{3pt}
\resizebox{0.5\textwidth}{!}{%
\begin{tabular}{llll}
\hline
\textbf{Difficulty}         & \textbf{Category} & \textbf{Description}   & \textbf{Count} \\ \hline
\multirow{4}{*}{Easy}   & E1          & Weather and Action     & 8,283           \\
                        & E2          & Existence              & 8,019           \\
                        & E3          & Counting               & 7,586           \\
                        & E4          & Attribute              & 7,570           \\ \hline
\multirow{3}{*}{Medium} & M1          & Spatial                & 7,670           \\
                        & M2          & Description            & 7,570           \\
                        & M3          & View Comparison        & 7,553           \\ \hline
\multirow{2}{*}{Hard}   & H1          & Risk Assessment        & 7,570           \\
                        & H2          & Navigational Landmarks & 7,570           \\ \hline
\end{tabular}%
}
\end{wraptable}

% we design 9 VQA categories with 3 difficult levels: \textit{easy}, \textit{medium}, and \textit{hard} difficulty. The detailed categories and distributions are shown in Table~\ref{tab:difficulty_levels}. 

\textbf{VQA Types.} To create \textbf{mmWalkVQA}, we designed a total of $9$ visual question categories, grouped into three difficult levels:\textit{easy}, \textit{medium}, and \textit{hard}, as shown in Table~\ref{tab:difficulty_levels}, with detailed categories and distributions.

\begin{figure}[h]
    \centering
    % \includesvg[width=1\textwidth,inkscapelatex=false]
    \includegraphics[width=1\textwidth]{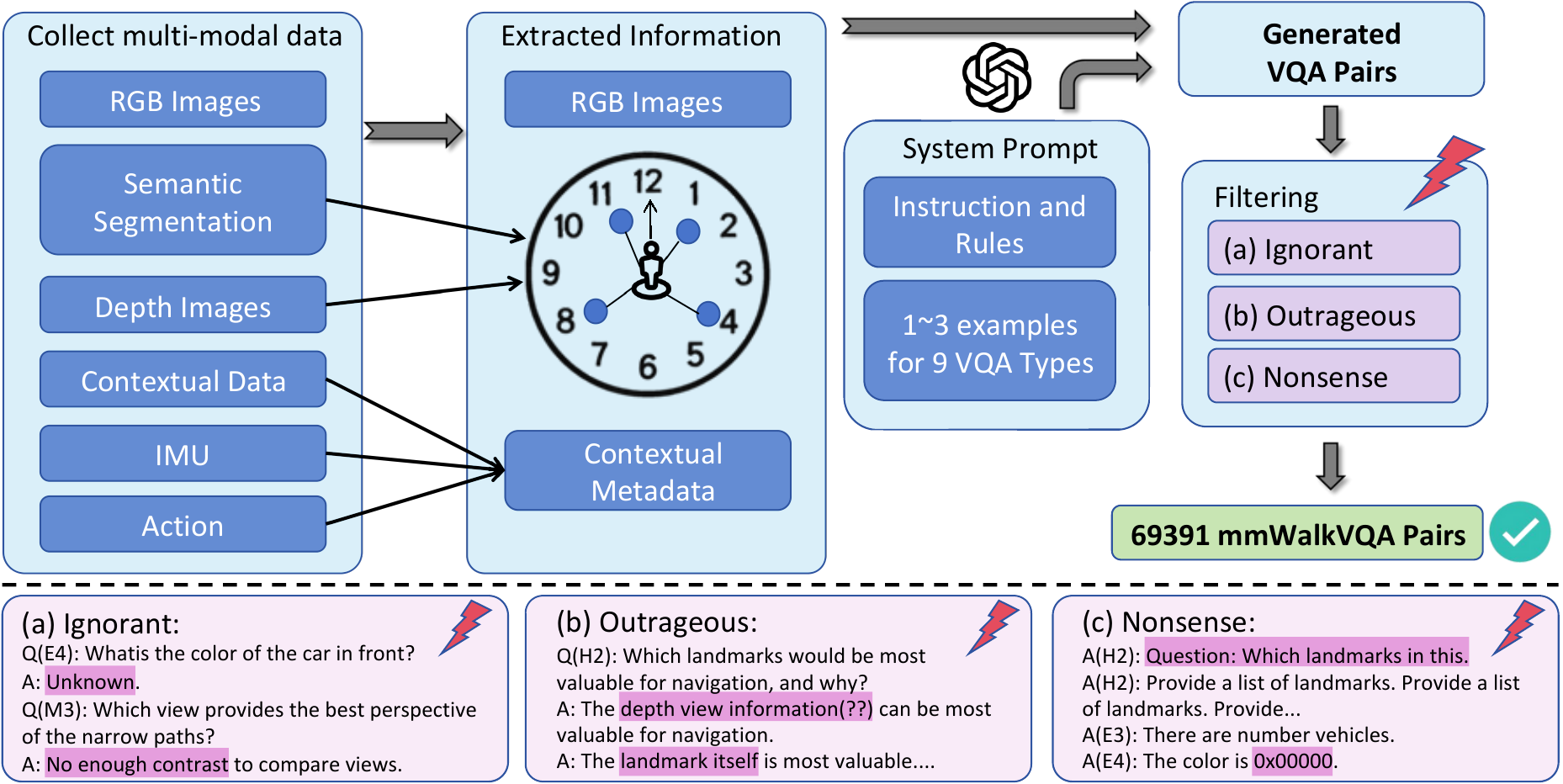}
    \caption{Workflow of VQA triplets generation.}
    \label{fig:VQAGen}
\end{figure}

\textbf{VQA Generation. } We randomly sample $7570$ frames among all trajectories for mmWalkVQA generation from the mmWalk dataset. We then used GPT-4o to generate the VQA pairs in batches. The overall workflow of VQA generation is shown in Figure~\ref{fig:VQAGen}.

In terms of scene information extraction, we kept the RGB images in all views, adhering to our multi-view concept, while translating the semantic segmentation and depth images into a list of strings describing object positions in a \textbf{BLV-friendly clockwise manner}~\cite{clock1,clock2,self}. Figure \ref{fig:VQAGen} illustrates the clockwise spatial descriptions from the ego perspective, where the blue nodes represent the semantic segments. The string lists were merged together with the contextual metadata. The system and instruction prompt are detailed in the Appendix~\ref{sec:qaappendix}. Additionally, we manually crafted $1{\sim}3$ example pairs for each VQA category. Those are fed into ChatGPT-4o to generate VQA pairs.  

\textbf{VQA Filtration and Validation. } To ensure high data quality, we first applied automatic filtering to remove low-quality VQA pairs by identifying manually defined keywords in questions and answers, such as \textit{Unknown, not possible, etc.}, as illustrated at the bottom of Figure~\ref{fig:VQAGen}. We iteratively refined the list of keywords until the filtering results were satisfactory. Additionally, manual corrections were applied to question-answer pairs when a quick fix was feasible. We also randomly sampled $270$ VQA pairs, and the authors manually examined them for quality. The manual exam result can be found in Table \ref{tab:VQAanalyse}, reporting average scores from the manual evaluation of $270$ sampled pairs across four criteria, where $5$ is the perfect score and $1$ is the lowest score. In particular, \textit{Answer Actionability} refers to whether the response provides users with clear, actionable guidance for decision-making.

This process resulted in a final set of \textbf{$69,391$} VQA triplets under $3$ difficulty levels and $9$ VQA types. We then split the dataset based on trajectory to ensure that the validation and testing sets do not contain similar scenes. Table \ref{tab:VQAanalyse} provides an overview of the training, validation, and test splits of the dataset. 

\begin{table}[H]
\centering
\caption{The data statistics and data splits of the established mmWalkVQA dataset. }
\label{tab:VQAanalyse}
\renewcommand\arraystretch{1.1}
\setlength{\tabcolsep}{4pt}
\resizebox{0.99\textwidth}{!}{%
\begin{tabular}{l|ccc|cc|cccc}
\hline
\multirow{2}{*}{\textbf{\begin{tabular}[c]{@{}c@{}}Difficulty\\ Level\end{tabular}}} & \multicolumn{3}{c|}{\textbf{Data Split}} & \multicolumn{2}{c|}{\textbf{Average Word Length}} & \multicolumn{4}{c}{\textbf{Quality Score}} \\
 & Train & Val & Test & Question & Answer & \begin{tabular}[c]{@{}c@{}}Question\\ Quality\end{tabular} & \begin{tabular}[c]{@{}c@{}}Answer\\ Correctness\end{tabular} & \begin{tabular}[c]{@{}c@{}}Answer\\ Actionability\end{tabular} & \begin{tabular}[c]{@{}c@{}}Answer\\ Fluency\end{tabular} \\ \hline
Easy & 25140 & 3214 & 3166 & 7.44 & 3.85 & 4.53 & 4.70 & 4.66 & 4.80 \\
Medium & 18201 & 2318 & 2249 & 8.69 & 22.74 & 4.71 & 4.87 & 4.84 & 4.91 \\
Hard & 12079 & 1537 & 1487 & 9.62 & 30.67 & 4.55 & 4.62 & 4.43 & 4.81 \\ \hline
\end{tabular}%
}
\end{table}

\textbf{Dataset Analysis. } 
We compare mmWalkVQA and five related datasets and benchmarks in Table \ref{tab:dataset_comparison}. Overall, mmWalkVQA features the largest size of VQA pairs and the most diverse Q types. Uniquely, it incorporates multi-view, panoramic, sequential trajectory, and egocentric images, yielding contextual richness and spatial accuracy of the data. These characteristics not only broaden the range of applicable scenarios but also enable the design of specialized VQA tasks tailored to the needs of the Blind and Low-Vision (BLV) community.
We also conducted a statistical analysis of the average word length, as presented in Table~\ref{tab:VQAanalyse}, and found that, as expected, longer questions and answers are associated with higher levels of difficulty.

\begin{table}[H]
    \centering
    \renewcommand{\arraystretch}{1.3}
    \caption{Comparison of datasets and benchmarks across multiple features with $\triangle$ indicates partially satisfied. MV: Multi-View. PA: Panoramic images. SD: Spatial or Depth. EC: Egocentric. BLV: BLV Guidance/Assistance. Seq: Sequential Trajectories. \#VQA: Number of VQA Pairs. \#VQA Type: Number of VQA types. FT: finetuned Evaluation. }
    \label{tab:dataset_comparison}
    \resizebox{\textwidth}{!}{
    \setlength{\tabcolsep}{10pt}
    \begin{tabular}{lcccccccccc}
        \toprule
        \textbf{Dataset} & Year & MV & PA & SD & EC & BLV & Seq & \#VQA & \#VQA Type & FT \\
        \midrule
        Vizwiz~\cite{gurari2018vizwiz}  &   CVPR'18    & $\times$ & $\times$ & $\times$ & $\checkmark$ & $\checkmark$ & $\times$ & 31,173 & 4 & $\checkmark$ \\
        SideGuide~\cite{sideguide}   & IROS'20  & $\times$ & $\times$ & $\checkmark$ & $\checkmark$ & $\checkmark$ & $\times$ & $\times$      & $\times$ & $\times$ \\
        SANPO~\cite{waghmare2023sanpo}   &   WACV'25    & $\checkmark$ & $\times$ & $\checkmark$ & $\checkmark$ & $\triangle$ & $\checkmark$ & $\times$      & $\times$ & $\times$ \\
        EgoTextVQA~\cite{zhou2025egotextVQA} & CVPR'25  & $\times$ & $\times$ & $\times$ & $\checkmark$ & $\times$ & $\checkmark$ & 7,064  & 5 & $\times$ \\
        GUIDEDOG~\cite{kim2025guidedog}   &  arXiv'2503  & $\times$ & $\times$ & $\checkmark$ & $\triangle$ & $\checkmark$ & $\checkmark$ & 818   & 2 & $\checkmark$ \\
        \rowcolor[gray]{.9} \multicolumn{2}{l}{\textbf{mmWalkVQA (ours)}}  & $\checkmark$ & $\checkmark$ & $\checkmark$ & $\checkmark$ & $\checkmark$ & $\checkmark$ & 69,391 & 9 & $\checkmark$ \\
        \bottomrule
    \end{tabular}
    }
\end{table}

%%%%%%%%% Experiments
\section{Model Benchmarking}
\label{sec:experiments}
\subsection{Baseline Models}
For benchmarking, we selected a number of open-source, multi-image input-enabled visual language models with language model sizes of 7${\sim}$8B, including \textbf{LLaVA One-Vision}~\cite{llava-OV}, \textbf{LLaVA Next}~\cite{liu2024llavanext}, \textbf{Qwen2VL}~\cite{wang2024qwen2vlenhancingvisionlanguagemodels}, \textbf{InternVL2}~\cite{chen2024internvl}, \textbf{Janus-Pro}~\cite{chen2025januspro}, and \textbf{Chameleon}~\cite{chameleonteam2025chameleonmixedmodalearlyfusionfoundation}.

\subsection{Evaluation Metric}
\label{subsec:score}
% With this mmWalkVQA content, we were able to detail a series of evaluation experiments in question answering tasks.

In our experiments, following \textit{LLM as-a-judge}~\cite{llmjudge1,llmjudge2}, we use GPT-4o-mini~\cite{openai2024gpt4o} to evaluate the similarity and correctness of the output answers generated by every model and the ground truth answers generated by GPT-4o, which are presupposed in our annotation and mentioned in Section~\ref{sec:dataset}. We designed a unified custom scoring prompt (detailed in Appendix~\ref{sec:qaappendix}) for evaluation of all models, which is input to GPT-4o-mini along with the questions, generated answers, and ground truth answers. The metric is a scaled score ($1{\sim}5$, with $5$ being the highest score). In order to visualize the differences in model performance, we normalized all scores to $0{\sim}100$ in all tables of experimental results. 
The data represented is the normalized scores $S_{normalized}$, which are given by the formula:
\[
S_{normalized} = \frac{1}{N_{samples}} \sum_{i=1}^{N_{samples}} \frac{s_i - 1}{4} \times 100\%,
\] 
where $s_i$ represents the original score of the $i$-th sample and $N_{samples}$ represents the number of all samples. Note that the data presented in all tables are accurate to two decimal places ($.2f$), which could lead to minor errors (${\leq}0.1$) in calculated average scores.

% \subsection{I: Basic Experiments}
\subsection{Overall Results}
Table \ref{tab:basic1} presents the performance of all models across each VQA category under both the zero-shot and 3-shot setups, accompanied by a composite average score. 

Overall, we found that all the models struggle on this task for all VQA categories with each model exhibiting strengths and weaknesses across different categories. These results indicate that the mmWalkVQA task presents substantial challenges. They also reveal inconsistencies in the capabilities of current open-source models. The result shows that InternVL2 performs the best among the zero-shot models, slightly outperforming LLaVA-Next and Qwen2VL. InternVL2 has a superior performance in the M3 category over other models, and a stable performance in all other categories. We thus chose InternVL2-8B for fine-tuning and evaluated 3-shot performance for the remaining models. Among the three-shot models, Qwen2VL-7B or Janus-Pro-7B consistently achieved the highest scores across all VQA categories. However, LLaVA-Next consistently ranked in the top three in most categories, resulting in the highest average score.

\begin{table}[H]
\vspace{-10pt}
\centering
\caption{Results on mmWalkVQA over all VQA categories, taking RGB panoramic image of all three views as input. %Scores are as stated in \ref{subsec:score}. 
The {\color[HTML]{158C1B}\textbf{best zero-shot model}} and the {\color[HTML]{3531FF} \textbf{ best 3-shot model}} for each VQA category and average score have been highlighted. The last column shows the improvement in scores at the 3-shot or finetuned model compared to the same zero-shot model.}
\label{tab:basic1}
\resizebox{\columnwidth}{!}{%
\begin{tabular}{@{}lllllllllll|cc@{}}
\toprule
\multicolumn{2}{l}{\textbf{Setting and Model}} & \textbf{E1} & \textbf{E2} & \textbf{E3} & \textbf{E4} & \textbf{M1} & \textbf{M2} & \textbf{M3} & \textbf{H1} & \textbf{H2} & \textbf{Average} & \textbf{Improved} \\ \midrule
\rowcolor[gray]{.9} \multicolumn{2}{l}{\textit{Zero-shot}} &  &  &  &  &  &  &  &  &  &  &  \\
 & Chameleon-7B & 25.78 & 20.64 & 16.01 & 15.39 & 7.45 & 19.85 & 29.17 & 28.19 & 27.36 & 21.14 &  \\
 & Janus-Pro-7B & 12.41 & {\color[HTML]{158C1B} \textbf{46.13}} & {\color[HTML]{158C1B} \textbf{33.40}} & 50.65 & 21.59 & {\color[HTML]{158C1B} \textbf{54.85}} & 31.01 & {\color[HTML]{158C1B} \textbf{54.30}} & {\color[HTML]{158C1B} \textbf{36.07}} & 37.82 &  \\
 & Qwen2VL-7B-Instruct & {\color[HTML]{158C1B} \textbf{84.01}} & 40.79 & 24.39 & 52.75 & 22.47 & 50.10 & 14.95 & 11.42 & 32.97 & 37.91 &  \\
 & LLaVA-NEXT-7B & 81.58 & 43.45 & 17.59 & {\color[HTML]{158C1B} \textbf{58.12}} & {\color[HTML]{158C1B} \textbf{24.71}} & 42.19 & 28.98 & 28.47 & 33.51 & 39.84 &  \\
 & LLaVA-Onevision-7B & 54.21 & 37.93 & 12.45 & 29.11 & 16.21 & 19.11 & 5.07 & 9.68 & 13.12 & 21.87 &  \\
 & InternVL2-8B & 77.56 & 42.66 & 31.22 & 53.12 & 17.03 & 46.51 & {\color[HTML]{158C1B} \textbf{53.26}} & 14.92 & 35.94 & {\color[HTML]{158C1B} \textbf{41.35}} &  \\ \midrule
\rowcolor[gray]{.9} \multicolumn{2}{l}{\textit{3-shot}} &  &  &  &  &  &  &  &  &  &  &  \\
 & Chameleon-7B & 27.91 & 22.67 & 15.46 & 16.64 & 13.32 & 8.65 & 33.01 & 27.89 & 36.84 & 22.48 & 1.34 \\
 & Janus-Pro-7B & 11.78 & {\color[HTML]{3531FF} \textbf{45.36}} & {\color[HTML]{3531FF} \textbf{33.61}} & 50.92 & {\color[HTML]{3531FF} \textbf{21.36}} & {\color[HTML]{3531FF} \textbf{55.08}} & 31.38 & {\color[HTML]{3531FF} \textbf{54.33}} & 35.97 & 37.75 & -0.07 \\
 & Qwen2VL-7B-Instruct & {\color[HTML]{3531FF} \textbf{84.01}} & 40.63 & 24.43 & {\color[HTML]{3531FF} \textbf{55.26}} & 15.72 & 45.83 & {\color[HTML]{3531FF} \textbf{45.33}} & 13.88 & {\color[HTML]{3531FF} \textbf{45.91}} & 41.89 & 3.98 \\
 & \begin{tabular}[c]{@{}l@{}}LLaVA-NEXT-7B\end{tabular} & 83.73 & 41.81 & 17.01 & 53.29 & 13.94 & 46.33 & 40.39 & 47.58 & 43.81 & {\color[HTML]{3531FF} \textbf{43.71}} & 3.87 \\
 & \begin{tabular}[c]{@{}l@{}}LLaVA-Onevision-7B\end{tabular} & 54.21 & 38.02 & 12.35 & 36.07 & 13.32 & 27.32 & 22.88 & 32.46 & 40.31 & 31.21 & 9.34 \\ \midrule
\rowcolor[gray]{.9} \multicolumn{2}{l}{\textit{finetuned}} &  &  &  &  &  &  &  &  &  &  &  \\
 & \textbf{InternVL2-8B} & \textbf{94.15} & \textbf{50.86} & \textbf{35.84} & \textbf{67.12} & \textbf{28.05} & \textbf{60.33} & \textbf{51.18} & \textbf{53.71} & \textbf{50.37} & \textbf{55.21} & \textbf{13.86} \\ \bottomrule
\end{tabular}%
}
\begin{flushleft}
Note: For the LLaVA family, the size of the language models we use remains 7-8B, with LLaVA-Onevision-qwen2-7B and LLaVA-Next-v1.6-mistral-7B used in the table above.
\end{flushleft}
\end{table}
% \footnotetext{Note that for the LLaVA family, the size of the language models we use remains 7-8B, with LLaVA-Onevision-qwen2-7B and LLaVA-Next-v1.6-mistral-7B used in the table above.}
\vspace{-10pt}

Among all tasks, M1 (Spatial) proved to be the most challenging category both for the zero-shot models and the 3-shot models, as M1 required a comprehensive spatial understanding, combining all the input modalities.

Across prompt settings, most models demonstrated improved performance under 3-shot learning compared to zero-shot. Among them, LLaVA-Onevision gained the largest improvement with more than $9.34\%$ improvement compared to zero-shot. In 0-shot cases, LLaVA-Onevision answered long texts in medium and hard questions with less relevance to key features, resulting in its low scores in M1-H2. With three examples provided, LLaVA-Onevision was better able to focus on the relevant aspects in its responses. Janus’s score decreased slightly, likely because it tended to mimic the example answers rather than reasoning from the input images while facing medium-level questions in our tasks.

When comparing model (\textit{i.e.}, InternVL2) with and without fine-tuning, we found that it improved its score by $13.86\%$, which proves the effectiveness of training on our proposed mmWalkVQA dataset. 

\subsection{Fine-Grained Analysis}
\textbf{Analysis of Different Walking Scenarios. } Table~\ref{tab:basic2} shows results of models across different Scenario categories. For each model, the highest-scoring scenario is highlighted in the table. The tasks associated with large open area (\textit{e.g.}, Plaza, Parking) generally yielded higher scores among many models. In contrast, scenarios such as Corner, Busstop, and Gasstation, being more intricate and requiring finer-grained understanding, led to lower performance.
\begin{table}[t]
% \scriptsize
\centering
\caption{Results on mmWalkVQA over walking scenarios, taking RGB panoramic image of all three views as input. The {\color[HTML]{158C1B}\textbf{scenario-wise best performance}} of each model (each row) has been highlighted.}
\label{tab:basic2}
\setlength{\tabcolsep}{10pt}
\resizebox{\columnwidth}{!}{%
\begin{tabular}{@{}lllllllll@{}}
\toprule
\multicolumn{2}{l}{\textbf{Setting and Model}} & \textbf{Corner} & \textbf{Cross} & \textbf{Busstop} & \textbf{Mixed} & \textbf{Plaza} & \textbf{Parking} & \textbf{Gasstation} \\ \midrule
\rowcolor[gray]{.9} \multicolumn{2}{l}{\textit{Zero-shot}} &  &  &  &  &  &  &  \\
 & Chameleon-7B & 18.66 & {\color[HTML]{158C1B} \textbf{24.36}} & 18.59 & 21.68 & 22.60 & 21.71 & 20.74 \\
 & Janus-Pro-7B & 36.50 & 31.29 & 32.64 & 42.37 & {\color[HTML]{158C1B} \textbf{45.35}} & 29.77 & 35.84 \\
 & Qwen2VL-7B-Instruct & 33.81 & 34.57 & 34.52 & 38.76 & {\color[HTML]{158C1B} \textbf{45.14}} & 39.91 & 34.09 \\
 & LLaVA-NEXT-7B & 37.65 & {\color[HTML]{158C1B} \textbf{41.12}} & 37.11 & 40.17 & 38.95 & 37.46 & 37.26 \\
 & LLaVA-Onevision2-7B & 21.36 & 26.40 & 15.48 & 24.81 & 26.54 & {\color[HTML]{158C1B} \textbf{28.55}} & 12.55 \\
 & InternVL2-8B & 38.26 & 41.95 & 41.01 & 42.37 & {\color[HTML]{158C1B} \textbf{45.08}} & 42.36 & 40.77 \\ \midrule
\rowcolor[gray]{.9} \multicolumn{2}{l}{\textit{3-shot}} &  &  &  &  &  &  &  \\
 & Chameleon-7B & 22.28 & 22.81 & 21.22 & 23.81 & 23.45 & {\color[HTML]{158C1B} \textbf{23.65}} & 19.27 \\
 & Janus-Pro-7B & 36.23 & 31.47 & 32.97 & {\color[HTML]{158C1B} \textbf{44.89}} & 42.07 & 30.07 & 35.75 \\
 & Qwen2VL-7B-Instruct & 39.31 & 38.40 & 37.99 & 42.93 & {\color[HTML]{158C1B} \textbf{48.36}} & 44.01 & 38.12 \\
 & LLaVA-NEXT-7B & {\color[HTML]{158C1B} \textbf{47.65}} & 45.02 & 42.54 & 42.04 & 42.03 & 45.11 & 43.84 \\
 & LLaVA-Onevision2-7B & 26.98 & 35.69 & 24.02 & {\color[HTML]{158C1B} \textbf{38.70}} & 32.98 & 33.61 & 24.24 \\ \midrule
\rowcolor[gray]{.9} \multicolumn{2}{l}{\textit{finetuned}} &  &  &  &  &  &  &  \\
 & InternVL2-8B & 53.67 & 51.94 & 54.25 & {\color[HTML]{158C1B} \textbf{59.49}} & 56.87 & 50.17 & 56.58 \\ \bottomrule
\end{tabular}
}
\vskip -3ex
\end{table}
% \vspace{-20pt}

% interestingly, some models gets better performance for single/double input under scenarios. % such as M1 for the single/double input gets a slight score boost.

\begin{table}[t]
\centering
\caption{Results of different input views. Full stands for full view with walker, dog, and drone. 
% Walker stands for Walker's view. Dog+ and Drone+ stand for Dog + Walker and Drone + Walker views. 
Scores {\color[HTML]{158C1B}\textbf{better than full view}} have been highlighted, otherwise worse.}
\label{tab:adv}
\resizebox{\columnwidth}{!}{%
\begin{tabular}{@{}ccccccccccccc@{}}
\toprule
\multicolumn{2}{c}{\textbf{Models and Inputs}} & \textbf{E1} & \textbf{E2} & \textbf{E3} & \textbf{E4} & \textbf{M1} & \textbf{M2} & \textbf{M3} & \textbf{H1} & \textbf{H2} & \textbf{\begin{tabular}[c]{@{}c@{}}Average \\ without M3\end{tabular}} & \textbf{\begin{tabular}[c]{@{}c@{}}$\Delta$ to \\ Full\end{tabular}} \\ \midrule
\rowcolor[gray]{.9} \multicolumn{2}{l}{\textit{InternVL2-8B (finetuned)}} & \multicolumn{1}{l}{} & \multicolumn{1}{l}{} & \multicolumn{1}{l}{} & \multicolumn{1}{l}{} & \multicolumn{1}{l}{} & \multicolumn{1}{l}{} & \multicolumn{1}{l}{} & \multicolumn{1}{l}{} & \multicolumn{1}{l}{} & \multicolumn{1}{l}{} & \multicolumn{1}{l}{} \\
\textbf{} & Full & 94.15 & 50.86 & 35.84 & 67.12 & 28.05 & 60.33 & / & 53.71 & 50.37 & 55.05 & / \\
 & Walker & 93.12 & 48.34 & {\color[HTML]{158C1B} \textbf{38.49}} & 62.70 & 27.89 & 59.02 & / & 49.63 & 46.33 & 53.19 & {\color{red}\textbf{-1.86}} \\
 & Walker+Dog & 93.40 & 50.86 & {\color[HTML]{158C1B} \textbf{37.35}} & {\color[HTML]{158C1B} \textbf{62.98}} & {\color[HTML]{158C1B} \textbf{28.71}} & 59.93 & / & 51.04 & 46.74 & 53.85 & {\color{red}\textbf{-1.20}} \\
 & Walker+Drone & 94.10 & 50.15 & 35.68 & {\color[HTML]{158C1B} \textbf{67.22}} & {\color[HTML]{158C1B} \textbf{28.61}} & {\color[HTML]{158C1B} \textbf{60.80}} & / & 49.19 & 46.91 & 54.08 & {\color{red}\textbf{-0.97}} \\ \midrule
\rowcolor[gray]{.9} \multicolumn{2}{l}{\textit{LLaVA-NEXT-7B (zero-shot)}} & \multicolumn{1}{l}{} & \multicolumn{1}{l}{} & \multicolumn{1}{l}{} & \multicolumn{1}{l}{} & \multicolumn{1}{l}{} & \multicolumn{1}{l}{} & \multicolumn{1}{l}{} & \multicolumn{1}{l}{} & \multicolumn{1}{l}{} & \multicolumn{1}{l}{} & \multicolumn{1}{l}{} \\
 & Full & 81.58 & 43.45 & 17.59 & 58.12 & 24.71 & 42.19 & / & 28.47 & 33.51 & 41.20 & / \\
 & Walker & 74.86 & 42.54 & 15.23 & 51.29 & {\color[HTML]{158C1B} \textbf{25.07}} & {\color[HTML]{158C1B} \textbf{43.61}} & / & 27.72 & {\color[HTML]{158C1B} \textbf{36.47}} & 39.59 & {\color{red}\textbf{-1.61}} \\
 & Walker+Dog & 49.31 & 42.72 & {\color[HTML]{158C1B} \textbf{19.31}} & 54.31 & 23.98 & {\color[HTML]{158C1B} \textbf{42.83}} & / & 22.83 & {\color[HTML]{158C1B} \textbf{35.26}} & 36.06 & {\color{red}\textbf{-5.14}} \\
 & Walker+Drone & 70.01 & {\color[HTML]{158C1B} \textbf{43.52}} & {\color[HTML]{158C1B} \textbf{17.91}} & 54.21 & {\color[HTML]{158C1B} \textbf{24.80}} & 41.89 & / & 22.11 & {\color[HTML]{158C1B} \textbf{34.25}} & 38.58 & {\color{red}\textbf{-2.62}} \\ \midrule
\rowcolor[gray]{.9} \multicolumn{2}{l}{\textit{Qwen2VL-7B (zero-shot)}} & \multicolumn{1}{l}{} & \multicolumn{1}{l}{} & \multicolumn{1}{l}{} & \multicolumn{1}{l}{} & \multicolumn{1}{l}{} & \multicolumn{1}{l}{} & \multicolumn{1}{l}{} & \multicolumn{1}{l}{} & \multicolumn{1}{l}{} & \multicolumn{1}{l}{} & \multicolumn{1}{l}{} \\
 & Full & 84.01 & 40.79 & 24.39 & 52.75 & 22.47 & 50.10 & / & 11.42 & 32.97 & 39.86 & / \\
 & Walker & 78.23 & {\color[HTML]{158C1B} \textbf{41.62}} & 23.90 & 48.34 & {\color[HTML]{158C1B} \textbf{24.02}} & {\color[HTML]{158C1B} \textbf{51.75}} & / & {\color[HTML]{158C1B} \textbf{12.20}} & {\color[HTML]{158C1B} \textbf{33.48}} & 39.19 & {\color{red}\textbf{-0.67}} \\
 & Walker+Dog & 78.63 & 39.50 & 22.79 & 49.52 & 21.56 & {\color[HTML]{158C1B} \textbf{52.36}} & / & 11.36 & 32.74 & 38.55 & {\color{red}\textbf{-1.31}} \\
 & Walker+Drone & 77.82 & {\color[HTML]{158C1B} \textbf{41.41}} & {\color[HTML]{158C1B} \textbf{26.27}} & 50.82 & {\color[HTML]{158C1B} \textbf{23.59}} & {\color[HTML]{158C1B} \textbf{50.40}} & / & 11.20 & 32.44 & 39.36 & {\color{red}\textbf{-0.50}} \\ \bottomrule
\end{tabular}%
}
\end{table}

\begin{figure}[h]
    \centering
    % \includesvg[width=1\textwidth,inkscapelatex=false]
    \includegraphics[width=1\textwidth]{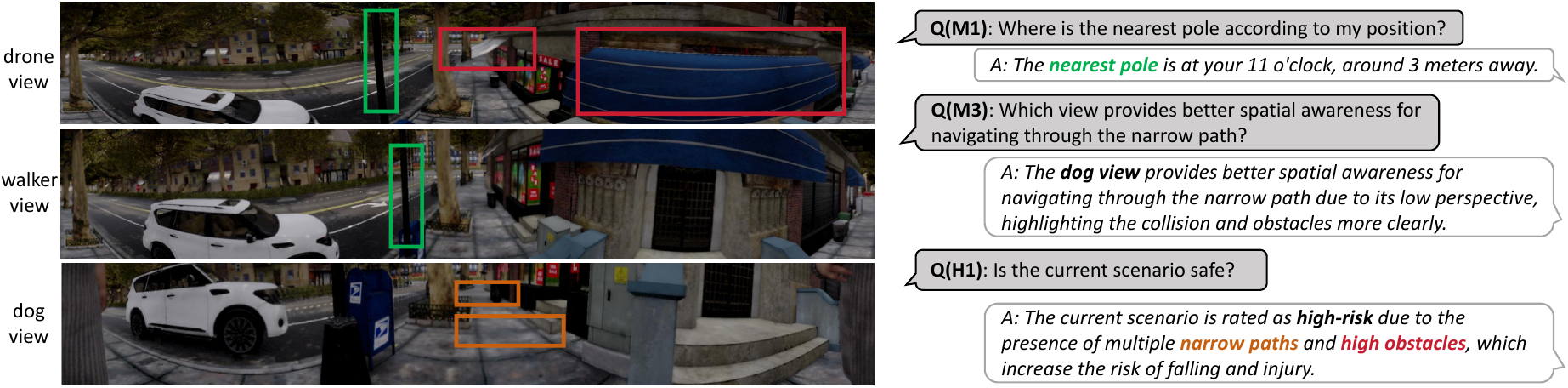}
    \caption{Qualitative examples of output from finetuned InternVL2-8B.}
    \label{fig:quali}
\end{figure}

\textbf{Analysis of Multi/Double/Single view Inputs.}
In real-world scenarios, BLV users often navigate independently, accompanied by a guide dog, or assisted by a drone, while full visual coverage may not always be accessible. Therefore, we conducted experiments using three input configurations: walker only, walker plus guide dog view, and walker plus drone view. The models we benchmarked are the top-3 models from our base experiment. Results are shown in Table~\ref{tab:adv}. As expected, full view generally outperforms single or dual views across all models on average. However, when examining specific tasks, we observed instances where single or dual views performed better, as highlighted in green, with an example shown in Figure~\ref{fig:quali}, where the question M1 can be answered easily by any views. Notably, for tasks like H1 (risk assessment), the drone view proved to be the least effective. This may be attributed to the drone's elevated perspective, which can miss ground-level risks such as uneven surfaces or narrow pathways. An illustrative example is shown in Figure~\ref{fig1:banner}, where the risk is only captured by the dog view.

\textbf{Reliability of LLM-as-a-judge} While LLM-based evaluation has become a widely adopted practice in recent works, the debate over whether LLM-as-a-judge is reliable remains ongoing. To further strengthen the reliability of the outcomes, we conducted an extra human rating study to directly quantify its consistency. Specifically, we randomly sampled 10 percent of VQA pairs from five model outputs, including Finetuned InternVL2-8B, Qwen2 in zero-shot and 3-shot settings, LLaVA-OneVision in the 3-shot setting, and LLaVA-Next in zero-shot. In total, $3,575$ VQA samples across all QA categories were rated by human annotators. We then calculated \textbf{Spearman’s $\boldsymbol{\rho}$}~\cite{spear} between the human ratings and GPT-4o evaluation scores.

Our results show that the finetuned model (InternVL2-8B) achieved a \textbf{Spearman’s $\boldsymbol{\rho}$} of 0.924, indicating very high agreement between human and GPT scores. Across all five models, the average \textbf{Spearman’s $\boldsymbol{\rho}$} reached 0.864, which demonstrates the overall reliability and consistency of our automatic evaluation pipeline. This fine-grained analysis confirms that our adoption of LLM-as-a-judge provides robust and scalable evaluation while maintaining high correlation with human judgment.

\textbf{Qualitative Examples.} In Figure~\ref{fig:quali}, we show three output examples of our finetuned \textit{InternVL2-8B} with full multi-view input. These examples show that our model is capable of providing reasonable answers for medium and hard VQA categories.
\subsection{Generalization on Real-World Dataset}
% To compensate that our dataset is based on a virtual simulator and to better show the contribution of our dataset, we deployed our finetuned model on related realistic datasets such as EgoTextVQA~\cite{zhou2025egotextVQA},
% following its original scoring metric to show the comparison and enhancement of our model. 

To present the generalization of our dataset, we evaluated the model finetuned on our dataset on the related realistic dataset, EgoTextVQA~\cite{zhou2025egotextVQA}, which contains outdoor scene VQs. 

Following the official scoring methodology from EgoTextVQA, we compared the performance of mmWalk finetuned \textit{InternVL2-8B} with the originally reported results. mmWalk finetuned the model with multi-image inputs, so we chose to use frames as input. The official EgoTextVQA paper published the \textit{InternVL2-8B} scores, including the score of each category with video input, and overall scores with frame input (first and second rows of the Table \ref{tab:egov}). Our model improves in almost all EgoTextVQA categories as well as in the overall score, as highlighted in the table.

\begin{table}[H]
\centering
\caption{Results of cross dataset evaluation following EgoTextVQA~\cite{zhou2025egotextVQA,egotextVQAgithub}. The score is rated by GPT-4o with accuracy/score (from $1$ to $5$, the higher the better). The {\color[HTML]{158C1B}\textbf{better performance}} compared with EgoTextVQA InternVL on frame input and video input has been highlighted. EgoTextVQA did not publish scores for each category for frame input; consequently, undisclosed scores are denoted as n.a. in the table.}
\label{tab:egov}
\renewcommand\arraystretch{1.2}
\resizebox{\columnwidth}{!}{%
\begin{tabular}{@{}llllllll@{}}
\toprule
\textbf{Model} & \textbf{Input} & \textbf{Location} & \textbf{Direction} & \textbf{Description} & \textbf{Intention Reasoning} & \textbf{Others} & \textbf{Overall} \\ \midrule
EgoTextVQA InternVL2-8B & Video & 15.8/1.4 & 21.9/1.7 & 14.8/1.0 & 14.5/1.2 & 13.6/1.3 & 16.4/1.3 \\
EgoTextVQA InternVL2-8B & Frame & \textit{n.a.} & \textit{n.a.} & \textit{n.a.} & \textit{n.a.} & \textit{n.a.} & 18.5/1.4 \\
\textbf{mmWalk-finetuned InternVL2-8B} & Frame & 11.82/1.59 & {\color[HTML]{158C1B} \textbf{22.58/2.05}} & {\color[HTML]{158C1B} \textbf{29.70/2.11}} & {\color[HTML]{158C1B} \textbf{23.12/1.95}} & {\color[HTML]{158C1B} \textbf{27.22/2.24}} & {\color[HTML]{158C1B} \textbf{21.55/1.92}} \\ \bottomrule
\end{tabular}%
}
\end{table}

%\textbf{III B. on Vizwiz}~\cite{gurari2018vizwiz} \textbf{for general BLV Assistance VQA}
%The Vizwiz dataset contains photographs taken by real BLV people and the related questions. 
%Deploying our model on the Vizwiz dataset helps to validate its ability to cope with blind people's question answers. Table~\ref{tab:vizw} shows the results of the mmWalk finetuned model compared with the Vizwiz experiments. Since Vizwiz has not yet published the ground-truth answers to the questions in the test split, we used the validation split instead, therefore, the results shown in the table are therefore for reference only rather than a full comparison. In terms of overall scores, our model still scores relatively well, and as expected, is not as good as the Vizwiz's own finetuned model based on their dataset.

\section{Conclusion and Discussion}
\label{sec:conclusion}
We present mmWalk, a multi-modal multi-view dataset for benchmarking walking assistance for individuals with Blindness or Low Vision (BLV). Our work addresses a critical gap in existing datasets by combining comprehensive multi-modal data with multiple viewpoints and a deliberate focus on corner cases and navigational landmarks specific to BLV users.
mmWalk uniquely features panoramic views, explicit annotation of BLV-relevant corner cases, and special navigational landmarks identified through ATmaps statistics. The accompanying mmWalkVQA benchmark enabled systematic evaluation of VLMs on BLV-relevant tasks, revealing significant performance gaps in state-of-the-art models, particularly in complex tasks like risk assessment and navigational landmark searching.

\section{Broader impacts, limitations, and future work.}
We expect our work to directly benefit BLV users by enhancing daily walking and navigation assistance and hope it can raise awareness among today's VLM developers for more inclusive models, thus having positive societal impacts. Additionally, it can support broader research communities in computer vision, including VQA, image captioning, pedestrian navigation, autonomous driving,  robotics, and embodied AI.

While promising, there are areas for improvement. Firstly, to comply with GDPR regulations, we collected the mmWalk dataset exclusively within a simulated environment, ensuring that no personal or sensitive information from real-world scenarios was captured, thereby maintaining data privacy and security. We also demonstrate that our dataset generalizes well to outdoor real-world VQA settings. 
% The mmWalk dataset was collected in simulation to ensure GDPR compliance and privacy, yet generalizes effectively to real-world outdoor VQA. 
However, a potential limitation is the risk of biased model behavior, as the training data are synthetic and may not fully capture the diversity of real-world BLV experiences, thus introducing potential negative societal impacts. Future work can address this by collecting data in real-world settings and easily adapting our proposed VQA generalization pipelines to mimic our tasks. Secondly, although using LLM-as-a-judge for answer evaluation has proven effective, it can also introduce biases. %, particularly favoring answers generated by the same model family (e.g., ChatGPT-4o as both judge and answer generator). 
Future work should investigate these biases and develop more stable evaluation metrics. 
% Beyond VQA, mmWalk could more comprehensively leverage its multi-modal data. 
Lastly, while mmWalk leverages multi-view and multi-modal features, some modalities, such as IMU data, sequential frames, and semantic labels, can be further explored. Expanding mmWalk to include these features more comprehensively could also open new avenues for tasks beyond VQA, such as image captioning and embodied AI training.

\section{Acknowledgment}
This work was supported in part by the Ministry of Science, Research and the Arts of Baden-W\"urttemberg (MWK) through the Cooperative Graduate School Accessibility through AI-based Assistive Technology (KATE) under Grant BW6-03, in part by funding from the pilot program Core-Informatics of the Helmholtz Association (HGF), in part by Karlsruhe House of Young Scientists (KHYS), and in part by the Helmholtz Association Initiative and Networking Fund on the HAICORE@KIT and HOREKA@KIT partition. 
This project is also supported in part by the National Natural Science Foundation of China under Grant No. 62473139, in part by the Hunan Provincial Research and Development Project (Grant No. 2025QK3019), and in part by the Open Research Project of the State Key Laboratory of Industrial Control Technology, China (Grant No. ICT2025B20). 

%%%%%%%%% REFERENCES

\clearpage
{
\small
\bibliographystyle{unsrt}
\bibliography{main}

\begin{thebibliography}{10}

\bibitem{WHO2023}
World~Health Organization.
\newblock Blindness and vision impairment, 2023.

\bibitem{naviReview1}
Mohamed~Dhiaeddine Messaoudi, Bob{-}Antoine~Jerry M{\'{e}}n{\'{e}}las, and Hamid Mcheick.
\newblock Review of navigation assistive tools and technologies for the visually impaired.
\newblock {\em Sensors}, 2022.

\bibitem{naviReview2}
A comprehensive review of navigation systems for visually impaired individuals.
\newblock {\em Heliyon}, 2024.

\bibitem{navigStreetCam}
Gaurav Jain, Basel Hindi, Zihao Zhang, Koushik Srinivasula, Mingyu Xie, Mahshid Ghasemi, Daniel Weiner, Sophie~Ana Paris, Xin Yi~Therese Xu, Michael~C. Malcolm, Mehmet~Kerem T{\"{u}}rkcan, Javad Ghaderi, Zoran Kostic, Gil Zussman, and Brian~A. Smith.
\newblock {StreetNav:} {Leveraging} street cameras to support precise outdoor navigation for blind pedestrians.
\newblock In {\em UIST}, 2024.

\bibitem{navigCV}
Praveen Nagil and Sumit~K. Mandal.
\newblock {DISHA:} {Low-energy} sparse transformer at edge for outdoor navigation for the visually impaired individuals.
\newblock In {\em ISLPED}, 2024.

\bibitem{landmark01}
Paulo Jorge~Sim{\~{o}}es Coelho, Hugo Fernandes, Ver{\'{o}}nica Vasconcelos, Paulo Coelho, Jo{\~{a}}o Barroso, and Leontios~J. Hadjileontiadis.
\newblock Landmarks detection to assist the navigation of visually impaired people.
\newblock In {\em HCI}, 2011.

\bibitem{landmark2}
Anna Wunderlich and Klaus Gramann.
\newblock Landmark-based navigation instructions improve incidental spatial knowledge acquisition in real-world environments.
\newblock {\em Journal of Environmental Psychology}, 2021.

\bibitem{ATMAPs}
ATmaps.eu.
\newblock Specification of symbols used on audio-tactile maps for individuals with blindness, 2023.

\bibitem{navSurvey}
Fatma El-Zahraa El-Taher, Luis Miralles-Pechu{\'a}n, Jane Courtney, Kristina Millar, Chantelle Smith, and Susan Mckeever.
\newblock A survey on outdoor navigation applications for people with visual impairments.
\newblock {\em IEEE Access}, 2023.

\bibitem{obstaInjury1}
Roberto Manduchi and Sri Kurniawan.
\newblock Mobility-related accidents experienced by people with visual impairment.
\newblock {\em Insight: Research and Practice in Visual Impairment and Blindness}, 2011.

\bibitem{corner+cornerAssist}
Manuel Zahn and Armaghan~Ahmad Khan.
\newblock Obstacle avoidance for blind people using a {3D} camera and a haptic feedback sleeve.
\newblock {\em arXiv preprint arXiv:2201.04453}, 2022.

\bibitem{corner+View}
Homa Kazemi, Mohammad Kamali, Reza Salehi, and Hossein Mobaraki.
\newblock Recognizing the viewpoint and experience of blind people in navigation and daily traffic.
\newblock {\em Function and Disability Journal}, 2023.

\bibitem{corner1}
Abbas Riazi, Fatemeh Riazi, Rezvan Yoosfi, and Fatemeh Bahmeei.
\newblock Outdoor difficulties experienced by a group of visually impaired iranian people.
\newblock {\em Journal of Current Ophthalmology}, 2016.

\bibitem{blindCross}
Blind and visually impaired pedestrians: What are their difficulties when crossing the street?, 2023.

\bibitem{carla}
Alexey Dosovitskiy, German Ros, Felipe Codevilla, Antonio Lopez, and Vladlen Koltun.
\newblock {CARLA}: {An} open urban driving simulator.
\newblock In {\em CoRL}, 2017.

\bibitem{openai2024gpt4o}
OpenAI.
\newblock {GPT-4o} system card.
\newblock {\em arXiv preprint arXiv:2410.21276}, 2024.

\bibitem{multiview1}
Ruiqi Cheng, Kaiwei Wang, Jian Bai, and Zhijie Xu.
\newblock {OpenMPR:} {Recognize} places using multimodal data for people with visual impairments.
\newblock {\em Measurement Science and Technology}, 2019.

\bibitem{multiview2}
Yang Di, Son~Lam Phung, and Abdesselam Bouzerdoum.
\newblock {MSSP:} {A} multi-view benchmark for street scene perception in assistive navigation.
\newblock In {\em IJCNN}, 2024.

\bibitem{drone1}
Pengfei Tong, Xuerong Yang, Yajun Yang, Wei Liu, and Peiyi Wu.
\newblock Multi-uav collaborative absolute vision positioning and navigation: A survey and discussion.
\newblock {\em Drones}, 7:261, 04 2023.

\bibitem{drone2}
Zhu Yun.
\newblock Research on aerial view point planning of drone based on multi-view.
\newblock {\em IOP Conference Series: Materials Science and Engineering}, 739:012055, 02 2020.

\bibitem{drone3}
Youzhi Liu, Fanglong Yao, Yuanchang Yue, Guangluan Xu, Xian Sun, and Kun Fu.
\newblock Navagent: Multi-scale urban street view fusion for uav embodied vision-and-language navigation, 2024.

\bibitem{drone4}
Muhammad~Yeasir Arafat, Muhammad~Morshed Alam, and Sangman Moh.
\newblock Vision-based navigation techniques for unmanned aerial vehicles: Review and challenges.
\newblock {\em Drones}, 7(2), 2023.

\bibitem{drone5}
Fabian Schilling, Fabrizio Schiano, and Dario Floreano.
\newblock Vision-based drone flocking in outdoor environments, 2021.

\bibitem{gurari2018vizwiz}
Danna Gurari, Qing Li, Abigale~J. Stangl, Anhong Guo, Chi Lin, Kristen Grauman, Jiebo Luo, and Jeffrey~P. Bigham.
\newblock {VizWiz} grand challenge: {Answering} visual questions from blind people.
\newblock In {\em CVPR}, 2018.

\bibitem{kim2025guidedog}
Junhyeok Kim, Jaewoo Park, Junhee Park, Sangeyl Lee, Jiwan Chung, Jisung Kim, Ji~Hoon Joung, and Youngjae Yu.
\newblock {GuideDog:} {A} real-world egocentric multimodal dataset for blind and low-vision accessibility-aware guidance.
\newblock {\em arXiv preprint arXiv:2503.12844}, 2025.

\bibitem{sideguide}
Kibaek Park, Youngtaek Oh, Soomin Ham, Kyungdon Joo, Hyokyoung Kim, Hyoyoung Kum, and In~So Kweon.
\newblock {SideGuide:} {A} large-scale sidewalk dataset for guiding impaired people.
\newblock In {\em IROS}, 2020.

\bibitem{tbrsd}
Junzhang Chen and Xiangzhi Bai.
\newblock Atmospheric transmission and thermal inertia induced blind road segmentation with a large-scale dataset tbrsd.
\newblock In {\em Proceedings of the IEEE/CVF International Conference on Computer Vision (ICCV)}, pages 1053--1063, October 2023.

\bibitem{xworld}
Jimuyang Zhang, Minglan Zheng, Matthew Boyd, and Eshed Ohn-Bar.
\newblock X-world: Accessibility, vision, and autonomy meet.
\newblock In {\em 2021 IEEE/CVF International Conference on Computer Vision (ICCV)}, pages 9742--9751, 2021.

\bibitem{waghmare2023sanpo}
Sagar~M. Waghmare, Kimberly Wilber, Dave Hawkey, Xuan Yang, Matthew Wilson, Stephanie Debats, Cattalyya Nuengsigkapian, Astuti Sharma, Lars Pandikow, Huisheng Wang, Hartwig Adam, and Mikhail Sirotenko.
\newblock {SANPO:} {A} scene understanding, accessibility and human navigation dataset.
\newblock In {\em WACV}, 2025.

\bibitem{ego4d}
Kristen Grauman, Andrew Westbury, Eugene Byrne, Zachary Chavis, Antonino Furnari, Rohit Girdhar, Jackson Hamburger, Hao Jiang, Miao Liu, Xingyu Liu, Miguel Martin, Tushar Nagarajan, Ilija Radosavovic, Santhosh~Kumar Ramakrishnan, Fiona Ryan, Jayant Sharma, Michael Wray, Mengmeng Xu, Eric~Zhongcong Xu, Chen Zhao, Siddhant Bansal, Dhruv Batra, Vincent Cartillier, Sean Crane, Tien Do, Morrie Doulaty, Akshay Erapalli, Christoph Feichtenhofer, Adriano Fragomeni, Qichen Fu, Abrham Gebreselasie, Cristina Gonzalez, James Hillis, Xuhua Huang, Yifei Huang, Wenqi Jia, Weslie Khoo, Jachym Kolar, Satwik Kottur, Anurag Kumar, Federico Landini, Chao Li, Yanghao Li, Zhenqiang Li, Karttikeya Mangalam, Raghava Modhugu, Jonathan Munro, Tullie Murrell, Takumi Nishiyasu, Will Price, Paola~Ruiz Puentes, Merey Ramazanova, Leda Sari, Kiran Somasundaram, Audrey Southerland, Yusuke Sugano, Ruijie Tao, Minh Vo, Yuchen Wang, Xindi Wu, Takuma Yagi, Ziwei Zhao, Yunyi Zhu, Pablo Arbelaez, David Crandall, Dima Damen, Giovanni~Maria
  Farinella, Christian Fuegen, Bernard Ghanem, Vamsi~Krishna Ithapu, C.~V. Jawahar, Hanbyul Joo, Kris Kitani, Haizhou Li, Richard Newcombe, Aude Oliva, Hyun~Soo Park, James~M. Rehg, Yoichi Sato, Jianbo Shi, Mike~Zheng Shou, Antonio Torralba, Lorenzo Torresani, Mingfei Yan, and Jitendra Malik.
\newblock {Ego4D:} {Around} the world in 3,000 hours of egocentric video.
\newblock In {\em CVPR}, 2022.

\bibitem{musohu}
Duc~M. Nguyen, Mohammad Nazeri, Amirreza Payandeh, Aniket Datar, and Xuesu Xiao.
\newblock Toward human-like social robot navigation: A large-scale, multi-modal, social human navigation dataset.
\newblock In {\em IROS}, 2023.

\bibitem{ma2025spatialllmcompound3dinformeddesign}
Wufei Ma, Luoxin Ye, Nessa McWeeney, Celso~M. de~Melo, Alan Yuille, and Jieneng Chen.
\newblock {SpatialLLM:} {A} compound {3D-informed} design towards spatially-intelligent large multimodal models.
\newblock In {\em CVPR}, 2025.

\bibitem{robotassist}
ByungOk Han, Woo han Yun, Beom-Su Seo, and Jaehong Kim.
\newblock Space-aware instruction tuning: Dataset and benchmark for guide dog robots assisting the visually impaired, 2025.

\bibitem{cross1}
Janet~M. Barlow, Billie~Louise Bentzen, and Tamara Bond.
\newblock Blind pedestrians and the changing technology and geometry of signalized intersections: Safety, orientation, and independence.
\newblock {\em Journal of Visual Impairment \& Blindness}, 2005.

\bibitem{cross2}
Bastian~J. Schroeder, Nagui~M. Rouphail, and Robert S.~Wall Emerson.
\newblock Exploratory analysis of crossing difficulties for blind and sighted pedestrians at channelized turn lanes.
\newblock {\em Transportation Research Record}, 2006.

\bibitem{uneven1}
Limin Zeng.
\newblock A survey: Outdoor mobility experiences by the visually impaired.
\newblock In {\em MuC (Workshopband)}, 2015.

\bibitem{barrier1}
Gabriel~Iluebe Okolo, Turke Althobaiti, and Naeem Ramzan.
\newblock Assistive systems for visually impaired persons: challenges and opportunities for navigation assistance.
\newblock {\em Sensors}, 2024.

\bibitem{barrier2}
Chengfeng Cai, Deer Liu, and Zhen Liu.
\newblock A spatial data model of blind outdoor navigation for path optimization.
\newblock {\em Humanities and Social Sciences Communications}, 2024.

\bibitem{entrance1}
Karst M.~P. Hoogsteen, Sarit Szpiro, Gabriel Kreiman, and Eli Peli.
\newblock Beyond the cane: Describing urban scenes to blind people for mobility tasks.
\newblock {\em ACM Transactions on Accessible Computing (TACCESS)}, 2022.

\bibitem{entrance2}
Bruno Berenguel-Baeta, Manuel Guerrero-Viu, A.~Nova, Jesus Bermudez-Cameo, Alejandro P{\'e}rez-Yus, and Josechu~J. Guerrero.
\newblock Floor extraction and door detection for visually impaired guidance.
\newblock In {\em ICARCV}, 2020.

\bibitem{entrance3}
Hilary Dalke, Alessio Corso, and Knights~Park Campus.
\newblock Making an entrance: colour, contrast and the design of entrances to homes of people with sight loss.
\newblock 2013.

\bibitem{high1}
Maxime Bleau, Samuel Par{\'e}, Isma{\"e}l Djerourou, Daniel~R. Chebat, Ron Kupers, and Maurice Ptito.
\newblock Blindness and the reliability of downwards sensors to avoid obstacles: A study with the {EyeCane}.
\newblock {\em Sensors}, 2021.

\bibitem{py360}
Brian~Pugh et. al.
\newblock py360convert, 2025.

\bibitem{clock1}
Mohammad~Moeen Valipoor and Ang{\'e}lica De~Antonio.
\newblock Recent trends in computer vision-driven scene understanding for vi/blind users: a systematic mapping.
\newblock {\em Universal Access in the Information Society}, 2023.

\bibitem{clock2}
Haikel Alhichri, Yakoub Bazi, and Naif Alajlan.
\newblock Assisting the visually impaired in multi-object scene description using {OWA-based} fusion of {CNN} models.
\newblock {\em Arabian Journal for Science and Engineering}, 2020.

\bibitem{self}
Kedi Ying, Mingzhe Tao, Ruize Dai, Ruiping Liu, Karin M{\"u}ller, Gerhard Jaworek, Jiaming Zhang, and Rainer Stiefelhagen.
\newblock {DishDetect:} {What’s} on my plate?--{Co-designing} a mobile app for clockwise food description.
\newblock In {\em CHI}, 2025.

\bibitem{zhou2025egotextVQA}
Sheng Zhou, Junbin Xiao, Qingyun Li, Yicong Li, Xun Yang, Dan Guo, Meng Wang, Tat-Seng Chua, and Angela Yao.
\newblock {EgoTextVQA:} {Towards} egocentric scene-text aware video question answering.
\newblock {\em arXiv preprint arXiv:2502.07411}, 2025.

\bibitem{llava-OV}
Bo~Li, Yuanhan Zhang, Dong Guo, Renrui Zhang, Feng Li, Hao Zhang, Kaichen Zhang, Peiyuan Zhang, Yanwei Li, Ziwei Liu, and Chunyuan Li.
\newblock {LLaVA-OneVision:} {Easy} visual task transfer.
\newblock {\em arXiv preprint arXiv:2408.03326}, 2024.

\bibitem{liu2024llavanext}
Haotian Liu, Chunyuan Li, Yuheng Li, Bo~Li, Yuanhan Zhang, Sheng Shen, and Yong~Jae Lee.
\newblock {LLaVA-NeXT:} {Improved} reasoning, {OCR}, and world knowledge, 2024.

\bibitem{wang2024qwen2vlenhancingvisionlanguagemodels}
Peng Wang, Shuai Bai, Sinan Tan, Shijie Wang, Zhihao Fan, Jinze Bai, Keqin Chen, Xuejing Liu, Jialin Wang, Wenbin Ge, Yang Fan, Kai Dang, Mengfei Du, Xuancheng Ren, Rui Men, Dayiheng Liu, Chang Zhou, Jingren Zhou, and Junyang Lin.
\newblock {Qwen2-VL:} {Enhancing} vision-language model's perception of the world at any resolution.
\newblock {\em arXiv preprint arXiv:2409.12191}, 2024.

\bibitem{chen2024internvl}
Zhe Chen, Jiannan Wu, Wenhai Wang, Weijie Su, Guo Chen, Sen Xing, Muyan Zhong, Qinglong Zhang, Xizhou Zhu, Lewei Lu, Bin Li, Ping Luo, Tong Lu, Yu~Qiao, and Jifeng Dai.
\newblock {Intern VL:} {Scaling} up vision foundation models and aligning for generic visual-linguistic tasks.
\newblock In {\em CVPR}, 2024.

\bibitem{chen2025januspro}
Xiaokang Chen, Zhiyu Wu, Xingchao Liu, Zizheng Pan, Wen Liu, Zhenda Xie, Xingkai Yu, and Chong Ruan.
\newblock {Janus-Pro:} {Unified} multimodal understanding and generation with data and model scaling.
\newblock {\em arXiv preprint arXiv:2501.17811}, 2025.

\bibitem{chameleonteam2025chameleonmixedmodalearlyfusionfoundation}
Chameleon Team.
\newblock Chameleon: Mixed-modal early-fusion foundation models.
\newblock {\em arXiv preprint arXiv:2405.09818}, 2024.

\bibitem{llmjudge1}
Haitao Li, Qian Dong, Junjie Chen, Huixue Su, Yujia Zhou, Qingyao Ai, Ziyi Ye, and Yiqun Liu.
\newblock {LLMs-as-judges:} {A} comprehensive survey on llm-based evaluation methods.
\newblock {\em arXiv preprint arXiv:2412.05579}, 2024.

\bibitem{llmjudge2}
Dawei Li, Renliang Sun, Yue Huang, Ming Zhong, Bohan Jiang, Jiawei Han, Xiangliang Zhang, Wei Wang, and Huan Liu.
\newblock Preference leakage: A contamination problem in {LLM-as-a-judge}.
\newblock {\em arXiv preprint arXiv:2502.01534}, 2025.

\bibitem{spear}
C.~Spearman.
\newblock The proof and measurement of association between two things.
\newblock {\em The American Journal of Psychology}, 15(1):72--101, 1904.

\bibitem{egotextVQAgithub}
Zhou Sheng.
\newblock {EgoTextVQA} {GitHub} repository.

\bibitem{pygame}
pygame.
\newblock pygame.

\end{thebibliography}
}

%%%%%%%%% Checklist
%%%%%%%%%%%%%%%%%%%%%%%%%%%%%%%%%%%%%%%%%%%%%%%%%%%%%%%%%%%%

\newpage
\section*{NeurIPS Paper Checklist}

\begin{enumerate}

\item {\bf Claims}
    \item[] Question: Do the main claims made in the abstract and introduction accurately reflect the paper's contributions and scope?
    \item[] Answer: \answerYes{} % Replace by \answerYes{}, \answerNo{}, or \answerNA{}.
    \item[] Justification: To present our contribution more intuitively, in the abstract and introduction we briefly cover all the motivations, dataset content, and results of the work. Detailed description of the dataset and benchmark can be found in the main paper.
    \item[] Guidelines:
    \begin{itemize}
        \item The answer NA means that the abstract and introduction do not include the claims made in the paper.
        \item The abstract and/or introduction should clearly state the claims made, including the contributions made in the paper and important assumptions and limitations. A No or NA answer to this question will not be perceived well by the reviewers. 
        \item The claims made should match theoretical and experimental results, and reflect how much the results can be expected to generalize to other settings. 
        \item It is fine to include aspirational goals as motivation as long as it is clear that these goals are not attained by the paper. 
    \end{itemize}

\item {\bf Limitations}
    \item[] Question: Does the paper discuss the limitations of the work performed by the authors?
    \item[] Answer: \answerYes{}
    \item[] Justification: The discussed limitations can be found at Section~\ref{sec:conclusion} of the main paper. 
    \item[] Guidelines:
    \begin{itemize}
        \item The answer NA means that the paper has no limitation while the answer No means that the paper has limitations, but those are not discussed in the paper. 
        \item The authors are encouraged to create a separate "Limitations" section in their paper.
        \item The paper should point out any strong assumptions and how robust the results are to violations of these assumptions (e.g., independence assumptions, noiseless settings, model well-specification, asymptotic approximations only holding locally). The authors should reflect on how these assumptions might be violated in practice and what the implications would be.
        \item The authors should reflect on the scope of the claims made, e.g., if the approach was only tested on a few datasets or with a few runs. In general, empirical results often depend on implicit assumptions, which should be articulated.
        \item The authors should reflect on the factors that influence the performance of the approach. For example, a facial recognition algorithm may perform poorly when image resolution is low or images are taken in low lighting. Or a speech-to-text system might not be used reliably to provide closed captions for online lectures because it fails to handle technical jargon.
        \item The authors should discuss the computational efficiency of the proposed algorithms and how they scale with dataset size.
        \item If applicable, the authors should discuss possible limitations of their approach to address problems of privacy and fairness.
        \item While the authors might fear that complete honesty about limitations might be used by reviewers as grounds for rejection, a worse outcome might be that reviewers discover limitations that aren't acknowledged in the paper. The authors should use their best judgment and recognize that individual actions in favor of transparency play an important role in developing norms that preserve the integrity of the community. Reviewers will be specifically instructed to not penalize honesty concerning limitations.
    \end{itemize}

\item {\bf Theory assumptions and proofs}
    \item[] Question: For each theoretical result, does the paper provide the full set of assumptions and a complete (and correct) proof?
    \item[] Answer: \answerNA{} % Replace by \answerYes{}, \answerNo{}, or \answerNA{}.
    \item[] Justification: This work is a dataset construction with benchmark experiments in the field of developing walking assistance. The models and methods are from previous work, therefore, it does not contain additional theoretical findings or results. 
    \item[] Guidelines:
    \begin{itemize}
        \item The answer NA means that the paper does not include theoretical results. 
        \item All the theorems, formulas, and proofs in the paper should be numbered and cross-referenced.
        \item All assumptions should be clearly stated or referenced in the statement of any theorems.
        \item The proofs can either appear in the main paper or the supplemental material, but if they appear in the supplemental material, the authors are encouraged to provide a short proof sketch to provide intuition. 
        \item Inversely, any informal proof provided in the core of the paper should be complemented by formal proofs provided in appendix or supplemental material.
        \item Theorems and Lemmas that the proof relies upon should be properly referenced. 
    \end{itemize}

    \item {\bf Experimental result reproducibility}
    \item[] Question: Does the paper fully disclose all the information needed to reproduce the main experimental results of the paper to the extent that it affects the main claims and/or conclusions of the paper (regardless of whether the code and data are provided or not)?
    \item[] Answer: \answerYes{} % Replace by \answerYes{}, \answerNo{}, or \answerNA{}.
    \item[] Justification: The inference and evaluation code used in the experiments, including the prompt and specific model name and size can be found. The code is publicly available for reproducing at: \url{https://github.com/KediYing/mmWalk}.
    \item[] Guidelines:
    \begin{itemize}
        \item The answer NA means that the paper does not include experiments.
        \item If the paper includes experiments, a No answer to this question will not be perceived well by the reviewers: Making the paper reproducible is important, regardless of whether the code and data are provided or not.
        \item If the contribution is a dataset and/or model, the authors should describe the steps taken to make their results reproducible or verifiable. 
        \item Depending on the contribution, reproducibility can be accomplished in various ways. For example, if the contribution is a novel architecture, describing the architecture fully might suffice, or if the contribution is a specific model and empirical evaluation, it may be necessary to either make it possible for others to replicate the model with the same dataset, or provide access to the model. In general. releasing code and data is often one good way to accomplish this, but reproducibility can also be provided via detailed instructions for how to replicate the results, access to a hosted model (e.g., in the case of a large language model), releasing of a model checkpoint, or other means that are appropriate to the research performed.
        \item While NeurIPS does not require releasing code, the conference does require all submissions to provide some reasonable avenue for reproducibility, which may depend on the nature of the contribution. For example
        \begin{enumerate}
            \item If the contribution is primarily a new algorithm, the paper should make it clear how to reproduce that algorithm.
            \item If the contribution is primarily a new model architecture, the paper should describe the architecture clearly and fully.
            \item If the contribution is a new model (e.g., a large language model), then there should either be a way to access this model for reproducing the results or a way to reproduce the model (e.g., with an open-source dataset or instructions for how to construct the dataset).
            \item We recognize that reproducibility may be tricky in some cases, in which case authors are welcome to describe the particular way they provide for reproducibility. In the case of closed-source models, it may be that access to the model is limited in some way (e.g., to registered users), but it should be possible for other researchers to have some path to reproducing or verifying the results.
        \end{enumerate}
    \end{itemize}

\item {\bf Open access to data and code}
    \item[] Question: Does the paper provide open access to the data and code, with sufficient instructions to faithfully reproduce the main experimental results, as described in supplemental material?
    \item[] Answer: \answerYes{} % Replace by \answerYes{}, \answerNo{}, or \answerNA{}.
    \item[] Justification: The dataset is publicly available at: \url{https://doi.org/10.7910/DVN/KKDXDK}. The code at: \url{https://github.com/KediYing/mmWalk}.
    \item[] Guidelines:
    \begin{itemize}
        \item The answer NA means that paper does not include experiments requiring code.
        \item Please see the NeurIPS code and data submission guidelines (\url{https://nips.cc/public/guides/CodeSubmissionPolicy}) for more details.
        \item While we encourage the release of code and data, we understand that this might not be possible, so “No” is an acceptable answer. Papers cannot be rejected simply for not including code, unless this is central to the contribution (e.g., for a new open-source benchmark).
        \item The instructions should contain the exact command and environment needed to run to reproduce the results. See the NeurIPS code and data submission guidelines (\url{https://nips.cc/public/guides/CodeSubmissionPolicy}) for more details.
        \item The authors should provide instructions on data access and preparation, including how to access the raw data, preprocessed data, intermediate data, and generated data, etc.
        \item The authors should provide scripts to reproduce all experimental results for the new proposed method and baselines. If only a subset of experiments are reproducible, they should state which ones are omitted from the script and why.
        \item At submission time, to preserve anonymity, the authors should release anonymized versions (if applicable).
        \item Providing as much information as possible in supplemental material (appended to the paper) is recommended, but including URLs to data and code is permitted.
    \end{itemize}

\item {\bf Experimental setting/details}
    \item[] Question: Does the paper specify all the training and test details (e.g., data splits, hyperparameters, how they were chosen, type of optimizer, etc.) necessary to understand the results?
    \item[] Answer: \answerYes{} % Replace by \answerYes{}, \answerNo{}, or \answerNA{}.
    \item[] Justification: Experimental setup can be found in Section \ref{sec:experiments}. The fine-tune model is publicly available in our code repository. Other details are in Appendix~\ref{experimentinfo}.
    \item[] Guidelines:
    \begin{itemize}
        \item The answer NA means that the paper does not include experiments.
        \item The experimental setting should be presented in the core of the paper to a level of detail that is necessary to appreciate the results and make sense of them.
        \item The full details can be provided either with the code, in appendix, or as supplemental material.
    \end{itemize}

\item {\bf Experiment statistical significance}
    \item[] Question: Does the paper report error bars suitably and correctly defined or other appropriate information about the statistical significance of the experiments?
    \item[] Answer: \answerYes{} % Replace by \answerYes{}, \answerNo{}, or \answerNA{}.
    \item[] Justification: The exact decimal point .2f error in the experimental results is presented in Section \ref{subsec:score} and other possible errors are discussed in Section \ref{sec:conclusion}.
    \item[] Guidelines:
    \begin{itemize}
        \item The answer NA means that the paper does not include experiments.
        \item The authors should answer "Yes" if the results are accompanied by error bars, confidence intervals, or statistical significance tests, at least for the experiments that support the main claims of the paper.
        \item The factors of variability that the error bars are capturing should be clearly stated (for example, train/test split, initialization, random drawing of some parameter, or overall run with given experimental conditions).
        \item The method for calculating the error bars should be explained (closed form formula, call to a library function, bootstrap, etc.)
        \item The assumptions made should be given (e.g., Normally distributed errors).
        \item It should be clear whether the error bar is the standard deviation or the standard error of the mean.
        \item It is OK to report 1-sigma error bars, but one should state it. The authors should preferably report a 2-sigma error bar than state that they have a 96\% CI, if the hypothesis of Normality of errors is not verified.
        \item For asymmetric distributions, the authors should be careful not to show in tables or figures symmetric error bars that would yield results that are out of range (e.g. negative error rates).
        \item If error bars are reported in tables or plots, The authors should explain in the text how they were calculated and reference the corresponding figures or tables in the text.
    \end{itemize}

\item {\bf Experiments compute resources}
    \item[] Question: For each experiment, does the paper provide sufficient information on the computer resources (type of compute workers, memory, time of execution) needed to reproduce the experiments?
    \item[] Answer: \answerYes{} % Replace by \answerYes{}, \answerNo{}, or \answerNA{}.
    \item[] Justification: The workspace computer resources for the experiment is described fully in Appendix~\ref{experimentinfo}. 
    \item[] Guidelines:
    \begin{itemize}
        \item The answer NA means that the paper does not include experiments.
        \item The paper should indicate the type of compute workers CPU or GPU, internal cluster, or cloud provider, including relevant memory and storage.
        \item The paper should provide the amount of compute required for each of the individual experimental runs as well as estimate the total compute. 
        \item The paper should disclose whether the full research project required more compute than the experiments reported in the paper (e.g., preliminary or failed experiments that didn't make it into the paper). 
    \end{itemize}
    
\item {\bf Code of ethics}
    \item[] Question: Does the research conducted in the paper conform, in every respect, with the NeurIPS Code of Ethics \url{https://neurips.cc/public/EthicsGuidelines}?
    \item[] Answer: \answerYes{} % Replace by \answerYes{}, \answerNo{}, or \answerNA{}.
    \item[] Justification: The dataset was captured in a virtual simulator and the work process did not violate any of the NeurIPS Code of Ethics articles.
    \item[] Guidelines:
    \begin{itemize}
        \item The answer NA means that the authors have not reviewed the NeurIPS Code of Ethics.
        \item If the authors answer No, they should explain the special circumstances that require a deviation from the Code of Ethics.
        \item The authors should make sure to preserve anonymity (e.g., if there is a special consideration due to laws or regulations in their jurisdiction).
    \end{itemize}

\item {\bf Broader impacts}
    \item[] Question: Does the paper discuss both potential positive societal impacts and negative societal impacts of the work performed?
    \item[] Answer: \answerYes{} % Replace by \answerYes{}, \answerNo{}, or \answerNA{}.
    \item[] Justification: We mention that the positive societal impact of this benchmark is enhancing the development of the walking and navigation assistance for people who are blind or have low vision (BVL). We also clarify that the potential negative impact is the risk of biased model behavior since the models are trained on synthetic and limited amount of data which does not fully cover the diversity of real-world walking experience.    
    \item[] Guidelines:
    \begin{itemize}
        \item The answer NA means that there is no societal impact of the work performed.
        \item If the authors answer NA or No, they should explain why their work has no societal impact or why the paper does not address societal impact.
        \item Examples of negative societal impacts include potential malicious or unintended uses (e.g., disinformation, generating fake profiles, surveillance), fairness considerations (e.g., deployment of technologies that could make decisions that unfairly impact specific groups), privacy considerations, and security considerations.
        \item The conference expects that many papers will be foundational research and not tied to particular applications, let alone deployments. However, if there is a direct path to any negative applications, the authors should point it out. For example, it is legitimate to point out that an improvement in the quality of generative models could be used to generate deepfakes for disinformation. On the other hand, it is not needed to point out that a generic algorithm for optimizing neural networks could enable people to train models that generate Deepfakes faster.
        \item The authors should consider possible harms that could arise when the technology is being used as intended and functioning correctly, harms that could arise when the technology is being used as intended but gives incorrect results, and harms following from (intentional or unintentional) misuse of the technology.
        \item If there are negative societal impacts, the authors could also discuss possible mitigation strategies (e.g., gated release of models, providing defenses in addition to attacks, mechanisms for monitoring misuse, mechanisms to monitor how a system learns from feedback over time, improving the efficiency and accessibility of ML).
    \end{itemize}
    
\item {\bf Safeguards}
    \item[] Question: Does the paper describe safeguards that have been put in place for responsible release of data or models that have a high risk for misuse (e.g., pretrained language models, image generators, or scraped datasets)?
    \item[] Answer: \answerNA{} % Replace by \answerYes{}, \answerNo{}, or \answerNA{}.
    \item[] Justification: The dataset for this work is based on the publicly available carla simulator, and the model is based on a publicly available language model with no additional risks arising from our work.
    \item[] Guidelines:
    \begin{itemize}
        \item The answer NA means that the paper poses no such risks.
        \item Released models that have a high risk for misuse or dual-use should be released with necessary safeguards to allow for controlled use of the model, for example by requiring that users adhere to usage guidelines or restrictions to access the model or implementing safety filters. 
        \item Datasets that have been scraped from the Internet could pose safety risks. The authors should describe how they avoided releasing unsafe images.
        \item We recognize that providing effective safeguards is challenging, and many papers do not require this, but we encourage authors to take this into account and make a best faith effort.
    \end{itemize}

\item {\bf Licenses for existing assets}
    \item[] Question: Are the creators or original owners of assets (e.g., code, data, models), used in the paper, properly credited and are the license and terms of use explicitly mentioned and properly respected?
    \item[] Answer: \answerYes{} % Replace by \answerYes{}, \answerNo{}, or \answerNA{}.
    \item[] Justification: The used open source assets are reasonably and properly cited in the references and all licenses are respected.
    \item[] Guidelines:
    \begin{itemize}
        \item The answer NA means that the paper does not use existing assets.
        \item The authors should cite the original paper that produced the code package or dataset.
        \item The authors should state which version of the asset is used and, if possible, include a URL.
        \item The name of the license (e.g., CC-BY 4.0) should be included for each asset.
        \item For scraped data from a particular source (e.g., website), the copyright and terms of service of that source should be provided.
        \item If assets are released, the license, copyright information, and terms of use in the package should be provided. For popular datasets, \url{paperswithcode.com/datasets} has curated licenses for some datasets. Their licensing guide can help determine the license of a dataset.
        \item For existing datasets that are re-packaged, both the original license and the license of the derived asset (if it has changed) should be provided.
        \item If this information is not available online, the authors are encouraged to reach out to the asset's creators.
    \end{itemize}

\item {\bf New assets}
    \item[] Question: Are new assets introduced in the paper well documented and is the documentation provided alongside the assets?
    \item[] Answer: \answerYes{} % Replace by \answerYes{}, \answerNo{}, or \answerNA{}.
    \item[] Justification: This work introduces a new dataset mmWalk and its QA benchmark mmWalkQA.
    \item[] Guidelines:
    \begin{itemize}
        \item The answer NA means that the paper does not release new assets.
        \item Researchers should communicate the details of the dataset/code/model as part of their submissions via structured templates. This includes details about training, license, limitations, etc. 
        \item The paper should discuss whether and how consent was obtained from people whose asset is used.
        \item At submission time, remember to anonymize your assets (if applicable). You can either create an anonymized URL or include an anonymized zip file.
    \end{itemize}

\item {\bf Crowdsourcing and research with human subjects}
    \item[] Question: For crowdsourcing experiments and research with human subjects, does the paper include the full text of instructions given to participants and screenshots, if applicable, as well as details about compensation (if any)? 
    \item[] Answer: \answerNA{} % Replace by \answerYes{}, \answerNo{}, or \answerNA{}.
    \item[] Justification: This work does not involve crowdsourcing nor research with human subjects.
    \item[] Guidelines:
    \begin{itemize}
        \item The answer NA means that the paper does not involve crowdsourcing nor research with human subjects.
        \item Including this information in the supplemental material is fine, but if the main contribution of the paper involves human subjects, then as much detail as possible should be included in the main paper. 
        \item According to the NeurIPS Code of Ethics, workers involved in data collection, curation, or other labor should be paid at least the minimum wage in the country of the data collector. 
    \end{itemize}

\item {\bf Institutional review board (IRB) approvals or equivalent for research with human subjects}
    \item[] Question: Does the paper describe potential risks incurred by study participants, whether such risks were disclosed to the subjects, and whether Institutional Review Board (IRB) approvals (or an equivalent approval/review based on the requirements of your country or institution) were obtained?
    \item[] Answer: \answerNA{} % Replace by \answerYes{}, \answerNo{}, or \answerNA{}.
    \item[] Justification: This work does not involve crowdsourcing nor research with human subjects.
    \item[] Guidelines:
    \begin{itemize}
        \item The answer NA means that the paper does not involve crowdsourcing nor research with human subjects.
        \item Depending on the country in which research is conducted, IRB approval (or equivalent) may be required for any human subjects research. If you obtained IRB approval, you should clearly state this in the paper. 
        \item We recognize that the procedures for this may vary significantly between institutions and locations, and we expect authors to adhere to the NeurIPS Code of Ethics and the guidelines for their institution. 
        \item For initial submissions, do not include any information that would break anonymity (if applicable), such as the institution conducting the review.
    \end{itemize}

\item {\bf Declaration of LLM usage}
    \item[] Question: Does the paper describe the usage of LLMs if it is an important, original, or non-standard component of the core methods in this research? Note that if the LLM is used only for writing, editing, or formatting purposes and does not impact the core methodology, scientific rigorousness, or originality of the research, declaration is not required.
    %this research? 
    \item[] Answer: \answerYes{} % Replace by \answerYes{}, \answerNo{}, or \answerNA{}.
    \item[] Justification: The usage of LLM are detailed in Section \ref{sec:dataset} and \ref{sec:experiments}.
    \item[] Guidelines:
    \begin{itemize}
        \item The answer NA means that the core method development in this research does not involve LLMs as any important, original, or non-standard components.
        \item Please refer to our LLM policy (\url{https://neurips.cc/Conferences/2025/LLM}) for what should or should not be described.
    \end{itemize}

\end{enumerate}

%%%%%%%%% Supplementary
\clearpage
% \setcounter{page}{1}
% \setcounter{section}{0}
% \renewcommand{\thesection}{\Alph{section}}

% \twocolumn[
% \centering
% \Large
% \textbf{mmWalk: Towards Inclusive Multimodal Walking Embodied AI} \\
% \vspace{1.0em} 
% \Large \textbf{(Supplementary Material)} \\
% \vspace{2.0em}
% ]

\appendix

\section{mmWalk Dataset Details}
\label{app1}
The mmWalk manual dataset collection is divided into three main parts, design, collection, and annotation. All code scripts are original and were collected on personal computers, and it is still under discussion whether the dataset collection scripts will be made public or not. Note that the collection requires Python 3.7 and a certain PC configuration to run Carla Simulator and pygame~\cite{pygame}. Please refer to the official Carla Simulator webpage~\cite{carla} for detailed requirements. To ensure full compliance with GDPR regulations, we collected the mmWalk dataset exclusively within a simulated environment. This approach eliminates the risk of capturing personal or sensitive information from real-world scenarios, thereby upholding data privacy and security standards.
\subsection{mmWalk}
\begin{figure*}[h]
    \centering
    \includegraphics[width=1\textwidth]{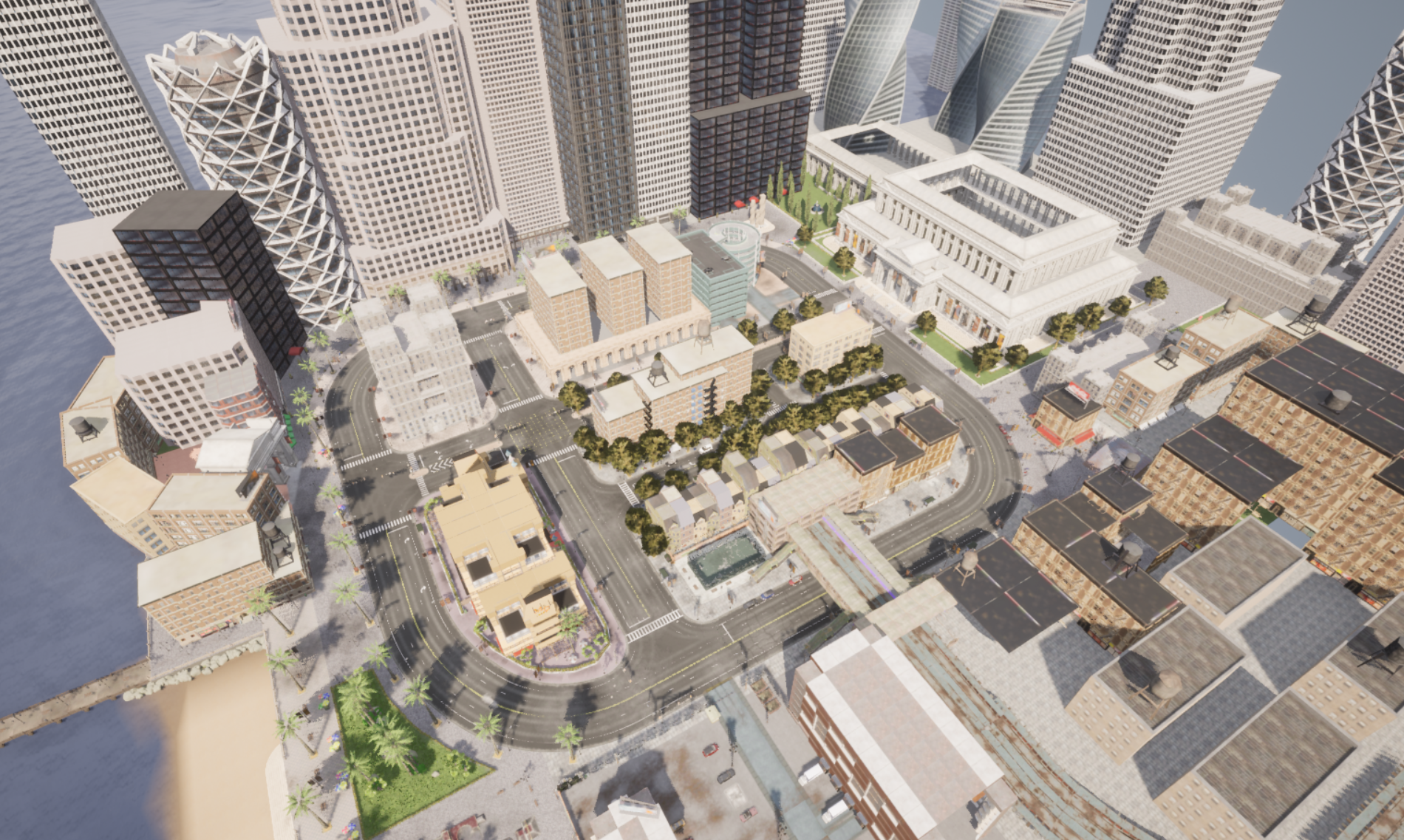}
    \caption{Overview of Carla map in Spectator perspective.}
    \label{fig:carla}
\end{figure*}

\textbf{Design.}
After running the Carla software, we can access a Carla's window as shown in Figure \ref{fig:carla} in the perspective of \textit{Spectator} under the player's control. We used a total of $9$ maps except these when designing the paths, mainly referring to the scenario and corner case mentioned in the main topic of the article, which are described in more detail in Tables \ref{tab:app_scenario} and \ref{tab:app_corner}.

\textbf{Collection.}
Depending on the requirements of our completed trajectory design, we moved the \textit{spectator} to the start point of the trajectory and run our collection script that integrates the selection of the weather settings in Figure~\ref{fig:carla-weather} and the panoramic image transformation~\cite{py360}. Once running, a new pygame window will appear that allows us to control the pedestrian's movement from the walker's first-person perspective, capturing keyboard inputs as action data and capturing original images in 4 directions for walker and dog view(left, front, right, back) and 6 directions for drone view(additional up and down) at a 2-second frame rate, meanwhile, dynamic moving hazards such as moving vehicles on the road will automatically spawn at random location nearby the ego pedestrain, controlled by the AI in the simulator with preset action and logic. The converted panoramic images are saved in a preset dictionary format as Figure~\ref{fig:dictionary}.

\textbf{Annotation.}
\label{app_annotation}
The metadata document can be found in the link to the dataset. The metadata information is all manually annotated, which includes the trajectory name, id, description, occurrence of corner cases, occurrence of important landmarks (Table~\ref{tab:app_landmark}), and the number of frames, and Table~\ref{tab:app_meta} provides the style of the metadata information. Of the $120$ trajectories, $103$ contained at least one type of corner cases, and $114$ contained at least one type of landmarks.

\subsection{mmWalkVQA}
\label{sec:qaappendix}
Table \ref{tab:app_qatype} provides details of the VQA categories and Table \ref{tab:qaexample} lists examples of each category given to GPT-4o \cite{openai2024gpt4o} along with the prompt in Figure~\ref{fig:app_qagen}. Figure \ref{fig:filter} shows some examples of what is handled in the filtering job. Figure \ref{fig:examplehard} gives good examples of generated pairs for hard-level VQA categories. Note that all the inputs for generating the VQA pairs are $3$ RGB images in different viewpoints and a spatial information matrix, here we only provide a certain image (or a certain part of it) for brevity.

\section{Experiment Details}
\label{experimentinfo}
\subsection{Models}
Table \ref{resource} shows the list of working environment resources and Table \ref{fig:finetune} shows the parameters of LoRA-finetune, the merged finetuned model and associated parameters are publicly available. Our finetune work took $10$ hours per $2$ Epochs, and for the inference in other experiments, the time consumed varied from $3{\sim}6$ hours with the performance of the model. Please check our submitted codes for more job-related information.

\subsection{Evaluation}
Figure \ref{fig:eval_gpt} illustrates the prompt used to score the model evaluation on a scale of $1{\sim}5$. After running the scoring, the script automatically gives a normalized score for each item based on the VQA category and the Scenario category, respectively, as well as an overall score with values accurate to .2f.

\newpage

\begin{table}[H]
\centering
\caption{Scenario and Detailed Description}
\label{tab:app_scenario}
\resizebox{\columnwidth}{!}{%
\begin{tabular}{@{}ll@{}}
\toprule
\textbf{Scenario} & \textbf{Description} \\ \midrule
Busstop & Trajectory starts near a bus stop, or the target location is heading to a bus stop. \\ \midrule
Gasstation & \begin{tabular}[c]{@{}l@{}}The entire trajectory uses the gas station or objects within the gas station \\ (\textit{e.g.}, parked cars, shops) as the target location or starting location.\end{tabular} \\ \midrule
Cross & The trajectory is based on the theme of crossing a road or multiple crossings. \\ \midrule
Parking & \begin{tabular}[c]{@{}l@{}}Trajectories that start or end with a core theme of car parks, with vehicles inside\\  the car parks or entrances and exits to the corresponding buildings.\end{tabular} \\ \midrule
Plaza & \begin{tabular}[c]{@{}l@{}}The theme of open squares and plaza, walking through a large square throughout \\ or looking for specific objects in the square (\textit{e.g.} benches, restaurants).\end{tabular} \\ \midrule
Mixed & Trajectories that mix two or more scenarios of all the above. \\ \midrule
Corner & \begin{tabular}[c]{@{}l@{}}Trajectories are designed purely on the basis of encountering and solving the \\ BLV Corner Cases.\end{tabular} \\ \bottomrule
\end{tabular}%
}
\end{table}

\begin{table}[H]
\centering
\caption{Corner Case with detailed description}
\label{tab:app_corner}
\resizebox{\columnwidth}{!}{%
\begin{tabular}{@{}ll@{}}
\toprule
\textbf{Corner Case} & \textbf{Detail Description} \\ \midrule
Entrance Locating & Searching correct path onto a small entrance of a house or a building. \\
High obstacles & High position obstacles may hurt head, face and influent the sensors \\
Deadend & Walking into a deadend, including turning around and walking back. \\
Uneven Road & \begin{tabular}[c]{@{}l@{}}Ground condition changes consisting of changes in topography, changes in ground materials, \\ standing water due to rain, broken glass bottle residue, going up and down stairs, cross a bridge \textit{etc.}\end{tabular} \\
Cross the Road in danger & Cross the road without traffic light or without pedestrian cross. \\
Cross the Road & Cross the road generally. \\
Barrier & \begin{tabular}[c]{@{}l@{}}Obstacles in the path blocking the way, making it necessary to make a diversion, including vehicles, \\ moving boxes, bikes or motorcycles occupying the path, terrain, large bushes, etc.\end{tabular} \\
Narrow Path & \begin{tabular}[c]{@{}l@{}}Walking through a narrow path that may be created by a natural scene (trees, plants, bushes) or \\ by vehicles, buildings, traffic lights, utility poles, etc.\end{tabular} \\ \bottomrule
\end{tabular}%
}
\end{table}

\newpage
\begin{figure}[H]
\centering
\begin{tcolorbox}[
  width=\textwidth,              % 让 box 和正文等宽
  colback=gray!5,
  colframe=black!70,
  arc=2mm,
  boxrule=0.4pt,
  left=5pt, right=5pt, top=3pt, bottom=5pt  % 内边距
]
\begin{lstlisting}[
  basicstyle=\rmfamily,
  breaklines=true,
  breakindent=0pt,
  columns=fullflexible
]
Sunny = {
  cloudiness=0, precipitation=0, precipitation_deposits=0,
  wind_intensity=30, wetness=0,
  sun_altitude_angle=75.0 }
Rainy = {
  cloudiness=80.0, precipitation=60.0, precipitation_deposits=40.0,
  wind_intensity=40.0, wetness=60.0,
  sun_altitude_angle=45.0 }
Foggy = {
  cloudiness=50.0, precipitation=0.0, precipitation_deposits=0.0,
  wind_intensity=30.0, wetness=0,
  sun_altitude_angle=60.0,
  fog_density=65.0, fog_distance=10.0, fog_falloff=1.0 }
Cloudy = {
  cloudiness=80.0, precipitation=0.0, precipitation_deposits=0.0,
  wind_intensity=50.0, wetness=0,
  sun_altitude_angle=65.0,
  fog_density=0.0 }
Night = {
  cloudiness=20.0, precipitation=0.0, precipitation_deposits=0.0,
  wind_intensity=15.0, wetness=0,
  sun_altitude_angle=-30.0,
  sun_azimuth_angle=270.0 }
\end{lstlisting}
\end{tcolorbox}
\caption{Weather configuration.}
\label{fig:carla-weather}
\end{figure}

\begin{figure}[H]
\centering
\begin{tcolorbox}[
  width=\textwidth,              % 让 box 和正文等宽
  colback=gray!5,
  colframe=black!70,
  arc=2mm,
  boxrule=0.4pt,
  left=5pt, right=5pt, top=3pt, bottom=5pt  % 内边距
]
\begin{lstlisting}[
  basicstyle=\rmfamily,
  breaklines=true,
  breakindent=0pt,
  columns=fullflexible
]
../Dataset
    /Busstop01
        /dog
            /rgb
                000001.png
                000002.png
                ......
            /semantic
                ......
            /depth
                ......
        /walker
            ... # Same as /dog Folder
        /drone
            ... # Same as /dog Folder
        /imu
            000001.txt
            000002.txt
            ......
        /action
            000001.txt
            000002.txt
            ......
    /Busstop02
    ......
\end{lstlisting}
\end{tcolorbox}
\caption{Dataset Dictionary.}
\label{fig:dictionary}
\end{figure}

\newpage
\begin{table}[]
\centering
\caption{Landmark List}
\label{tab:app_landmark}
\resizebox{\columnwidth}{!}{%
\begin{tabular}{@{}lll@{}}
\toprule
\textbf{id} & \textbf{content} & \textbf{comment} \\ \midrule
1 & Traffic Lights &  \\
2 & Bus Stop &  \\
3 & Entrance or Exit & Take in count only if the exit or entrance are preset as spawn or goal \\
4 & Stairs &  \\
5 & Square & including foundation, sightseeing square, square front of the church \textit{etc.} \\
6 & Pedestrian Cross & annotated only if the pedestrian cross the road \\
7 & Garbage Bin & small garbage bin \\
8 & Dumpster & large garbage dumpster \\
9 & Gate & large gate of parking area, construction site \textit{etc.} \\
10 & Bench &  \\
11 & Motocycle/Bycicle &  \\
12 & Poles & including streetlights and electric poles \\
13 & Postbox/Mailbox &  \\
14 & Map Board &  \\
15 & High voltage box & marked as dangerous for risk assessment \\
16 & Manhole Cover & marked as dangerous for risk assessment \\
17 & Roadside Stall & including Food stalls, newsstands, vending machines \\
18 & Money ATM &  \\ \bottomrule
\end{tabular}%
}
\end{table}

\begin{table}[H]
\centering
\caption{Example Metadata of $4$ different trajectories. / indicates no special landmark annotated. The same number in ID indicates the same path of these trajectories, yet opposite direction or different weather for comparison.}
\label{tab:app_meta}
\resizebox{\columnwidth}{!}{%
\begin{tabular}{@{}lllllll@{}}
\toprule
\textbf{Scenario} & \textbf{ID} & \textbf{Trajectory Description} & \textbf{Appeared Corner Case} & \textbf{Landmark id} & \textbf{Frame} & \textbf{Weather} \\ \midrule
Busstop & 14 & \begin{tabular}[c]{@{}l@{}}From bus station to home at night, \\ walking alongside a narrow sidewalk\end{tabular} & Narrow Path, Entrance Locating & 2,3,7,12,15 & 551 & Night \\ \midrule
Corner & 20A & \begin{tabular}[c]{@{}l@{}}Walking passby a Barrier on sidewalk, \\ go through multiple narrow path and \\ walk into a yard through uneven road.\\ (direction A)\end{tabular} & Narrow Path, Barrier, Uneven Road & / & 296 & Foggy \\
Corner & 20B & \begin{tabular}[c]{@{}l@{}}Walking from a yard through uneven road, \\ go through multiple narrow path and walk \\ passby a Barrier on sidewalk.\\ (direction B)\end{tabular} & Narrow Path, Barrier, Uneven Road & / & 271 & Foggy \\ \midrule
Mixed & 20 & \begin{tabular}[c]{@{}l@{}}Climb up stairs, cross 2 road continuously in \\ one intersection to reach the bus stop\end{tabular} & Uneven Road, Cross the road & 1,2,4,6,12 & 1249 & Sunny \\ \bottomrule
\end{tabular}%
}
\end{table}

\begin{table}[H]
\centering
\caption{Detailed Information of VQA Types}
\label{tab:app_qatype}
\resizebox{\columnwidth}{!}{%
\begin{tabular}{@{}lll@{}}
\toprule
\textbf{ID} & \textbf{Content} & \textbf{Type} \\ \midrule
E1 & Weather \& Action & query action and weather \\
E2 & Existence & query existence of one landmark, corner case or objects \\
E3 & Counting & counting landmarks or objects \\
E4 & Attribute & query attributes based on rgb-images \\
M1 & Spatial & \begin{tabular}[c]{@{}l@{}}query relative spatial information based on \\ rgb image and translated spatial description\end{tabular} \\
M2 & Description & full description of current scene \\
M3 & View Comparison & \begin{tabular}[c]{@{}l@{}}compare different views, in concern \\ corner cases or landmarks\end{tabular} \\
H1 & Risk Assessment & assess the risk level of current situation \\
H2 & Navigational Landmarks & evaluate landmarks about navigational value \\ \bottomrule
\end{tabular}%
}
\end{table}

\newpage
\begin{figure}[H]
\centering
\begin{tcolorbox}[
  width=\textwidth,              % 让 box 和正文等宽
  colback=gray!5,
  colframe=black!70,
  arc=2mm,
  boxrule=0.4pt,
  left=1pt, right=1pt, top=1pt, bottom=5pt  % 内边距
]
\begin{lstlisting}[
  basicstyle=\rmfamily,
  breaklines=true,
  breakindent=0pt,
  columns=fullflexible
]
SYSTEM_MESSAGE = "Generate 15 QA pairs based on multi-view scene graph and json file represented. A FRAME represents one frame of one trajectory, which contains one json file that describes this certain frame and path of 3 RGB images under three different views(dog, walker and drone), along with text descriptions of transformed semantic and depth information. Generate Question Answer pairs for each FRAME which should cover all 9 QA Types.(At least 1 pair for each type) The Question Answer pairs must follow the instruction and rules below, taking the given sample as example."

INSTRUCTION_CONTENT = "
Rules:
1.Only describe clear information in the images - do not fabricate or invent in the answers.
2.Base all answers only on what is actually visible in the provided rgb images and stated in the json data. Do not make assumptions or invent details.
3.All Position information must be described in clockwise manner.(Instead of left/right, describe exact clockwise location such as 'your 3 o'clock')
Instructions:
Easy Level QA: QA pairs that query the basic information in the json file or single image, the answer can be completely verified by the ground truth. Consider the questioner is the pedestrain in the first-person perspective of every scene, use "my surrounding" or current environment instead of "scene" in question.
-Type E1- General Query: the simple query questions about single feature(weather, action)), the answer should be concise and in several words or at most one sentence.
-Type E2- Existence: Query existence of specific objects or corner cases or landmarks in the current rgb image, the answer should be Yes or No with at most one sentence for extension, don't mention views in question and answers. '
-Type E3- Counting: Inquire the exact number of features, using only walker view, don't mention views in question and answers.
-Type E4- Attribute: simple query questions based on the RGB Image(color,shape,size,texture), the answer should be at most one sentence.
Middle Level QA: QA pairs that contain multiple views or multiple images in concern, the answer stems mainly from the combined ground truth feature information. The answer can be partially verified. Consider the questioner as an analyst instead of the pedestrain.
-Type M1- Spatial: Query about the spatial information (distance,position) between multiple objects.
-Type M2- Description: Describing the scene with all(or many) features depending on the question. Answers must be completely based on the information in multiple features and be at most 3 sentences, depending on the complexity of the scene, rely more on depth information.
-Type M3- Comparison:Compare different three views(dog/walker/drone), answers must be completely relevant to the enquiry and give more detailed reasons only based on ground truth. Answers must be at most 2 sentences, refer the appeared corner case in answer in concrete.
Hard Level QA: QA pairs that based on multiple features and multiple frames but require further merging, processing, analyzing and expanding. The question is on the abstract and summarizing perspective. The answer stems mainly from the processed ground truth information and image, refer the appeared corner case if there is any. The answer must be at most 3 sentences long.
-Type H1- Risk Assessment: Calculate the Risk of the scene based the information. The answer should give the final score in low/moderate/high risk based on image, concern weather, corner cases and risky landmarks and brief explanation in few sentences.
-Type H2- Landmarks and Navigation: Evaluate landmarks in the surroundings based on navigational value, and make an overall evaluation based on the size, distance, and number of landmarks, add concrete but concise reason based on image, not only saying that landmark has fixed position."

IMAGE_DESCRIPTION = "These images show the current scene from three camera positions (walker at eye level, drone from above, dog from low position). The RGB images show actual colors. Instead of depth and semantic images, we provide text descriptions of objects with their clock positions and approximate distances."
\end{lstlisting}
\end{tcolorbox}
\caption{VQA generation prompt.}
\label{fig:app_qagen}
\end{figure}

\newpage

\begin{table}[H]
\centering
\caption{Given examples of each VQA category}
\label{tab:qaexample}
\resizebox{\columnwidth}{!}{%
\begin{tabular}{@{}lll@{}}
\toprule
\textbf{ID} & \textbf{Q} & \textbf{A} \\ \midrule
E1-1 & What is the weather now? & Rainy. \\ \midrule
E1-2 & What is the current action? & Turning Left. \\ \midrule
E2-1 & Is there any benches in the scene? & No. \\ \midrule
E2-2 & Does the scene contain any corner cases for blind? & Yes, the scene contains multiple barriers and one narrow path. \\ \midrule
E3 & How many vehicles are there in the parking area? & 3 \\ \midrule
E4-1 & What is the color of the car in gas station? & Black \\ \midrule
E4-2 & How large is the Barrier in front of me? & \begin{tabular}[c]{@{}l@{}}There are some large cardboard boxes stacked on top of each other\\ on the pavement, 1 meter high and 1 meter wide.\end{tabular} \\ \midrule
M1-1 & \begin{tabular}[c]{@{}l@{}}What is the relative position of the sidewalk to \\ the bus stop?\end{tabular} & \begin{tabular}[c]{@{}l@{}}The bus stop is located on only one sidewalk, you are currently \\ standing on the side with bus stop.\end{tabular} \\ \midrule
M1-2 & How far is the intersection? & The intersection is around 20 meters far from you. \\ \midrule
M1-3 & Where is the bus stop according to my position? & The bus stop is 12 meters far at your 4 o'clock. \\ \midrule
M2 & Can you describe the scene & \begin{tabular}[c]{@{}l@{}}You are currently standing in the middle of a square, there are \\ few roadside stall nearby. There are multiple tables on your 2 o'clock. \\ There is a bus stop 16 meters far at your 7 o'clock.\end{tabular} \\ \midrule
M3 & \begin{tabular}[c]{@{}l@{}}Which view gives more information about \\ the dangerous environment?\end{tabular} & \begin{tabular}[c]{@{}l@{}}You are about to cross multiple narrow paths, the obstacles are clearer \\ in the 'dog' view. Whereas the drone view is partially obscured by the \\ foliage as the high obstacle. The dog view therefore provides the most \\ information to ensure your safety.\end{tabular} \\ \midrule
H1 & Is the current scenario safe? & \begin{tabular}[c]{@{}l@{}}The current scenario is rated generally as high-risk due to foggy weather \\ and the need to cross the road ahead without street lights. Foggy days \\ can cause drivers of vehicles to have a lower visible distance and a greater \\ potential threat when crossing the road.\end{tabular} \\ \midrule
H2 & \begin{tabular}[c]{@{}l@{}}Which landmarks in this scene would be most \\ valuable for a blind pedestrian's navigation, why?\end{tabular} & \begin{tabular}[c]{@{}l@{}}The traffic light at 3 meters straight ahead in the direction of your 11 o'clock \\ serves as the most valuable landmark due to its size and position. The bus \\ stop 8 meters far at your 2 o'clock offers secondary value with its shelter \\ structure, providing tactile and spatial reference. The storefront signs on \\ the left, while visible, have lower navigational value due to potential changes.\end{tabular} \\ \bottomrule
\end{tabular}%
}
\end{table}

\newpage

\begin{figure}[H]
    \centering
    % \includesvg[width=1\textwidth,inkscapelatex=false]
    \includegraphics[width=1\textwidth]{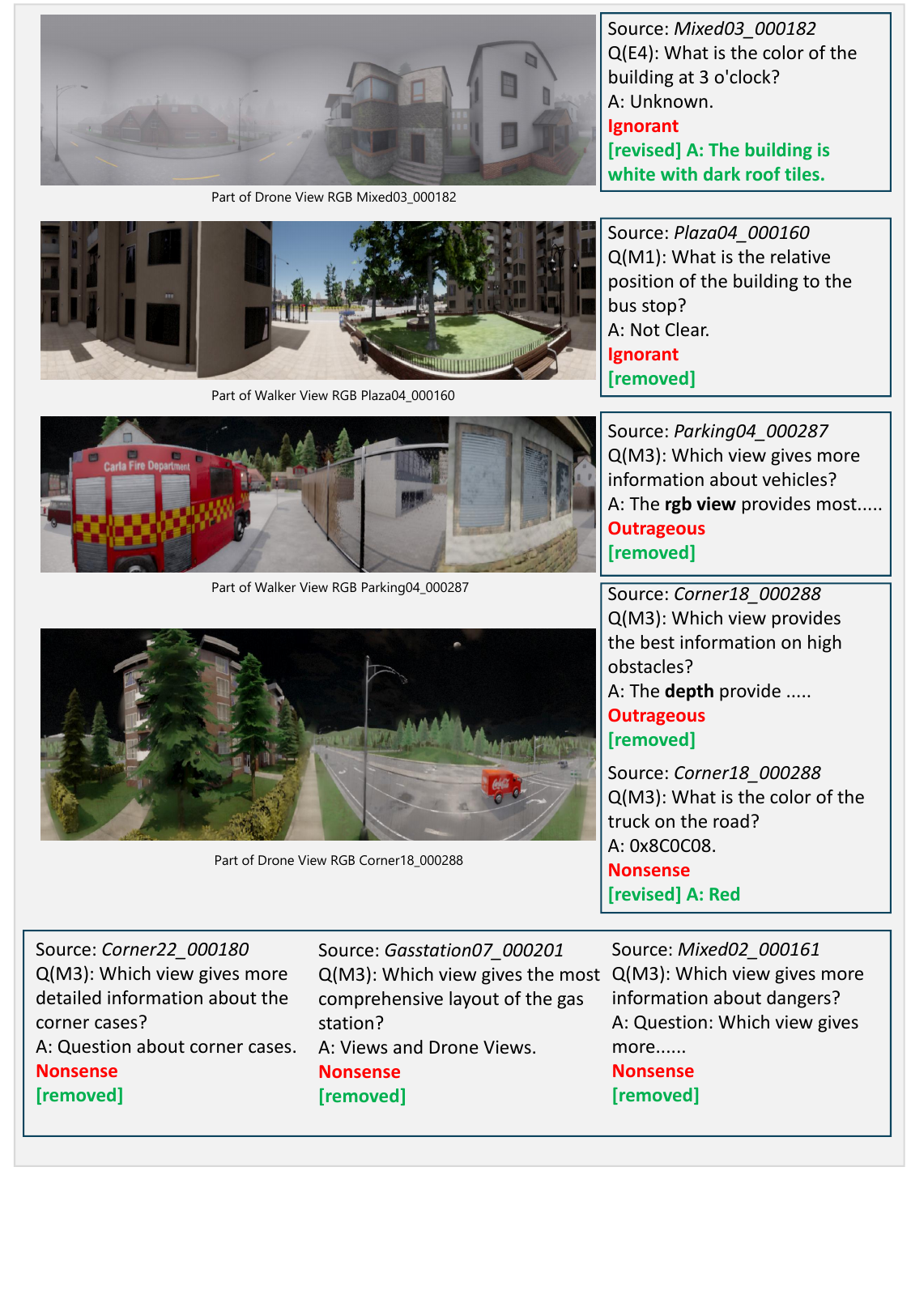}
    \caption{Examples of filtering work}
    \label{fig:filter}
\end{figure}

\newpage
\begin{figure}[H]
    \centering
    % \includesvg[width=1\textwidth,inkscapelatex=false]
    \includegraphics[width=1\textwidth]{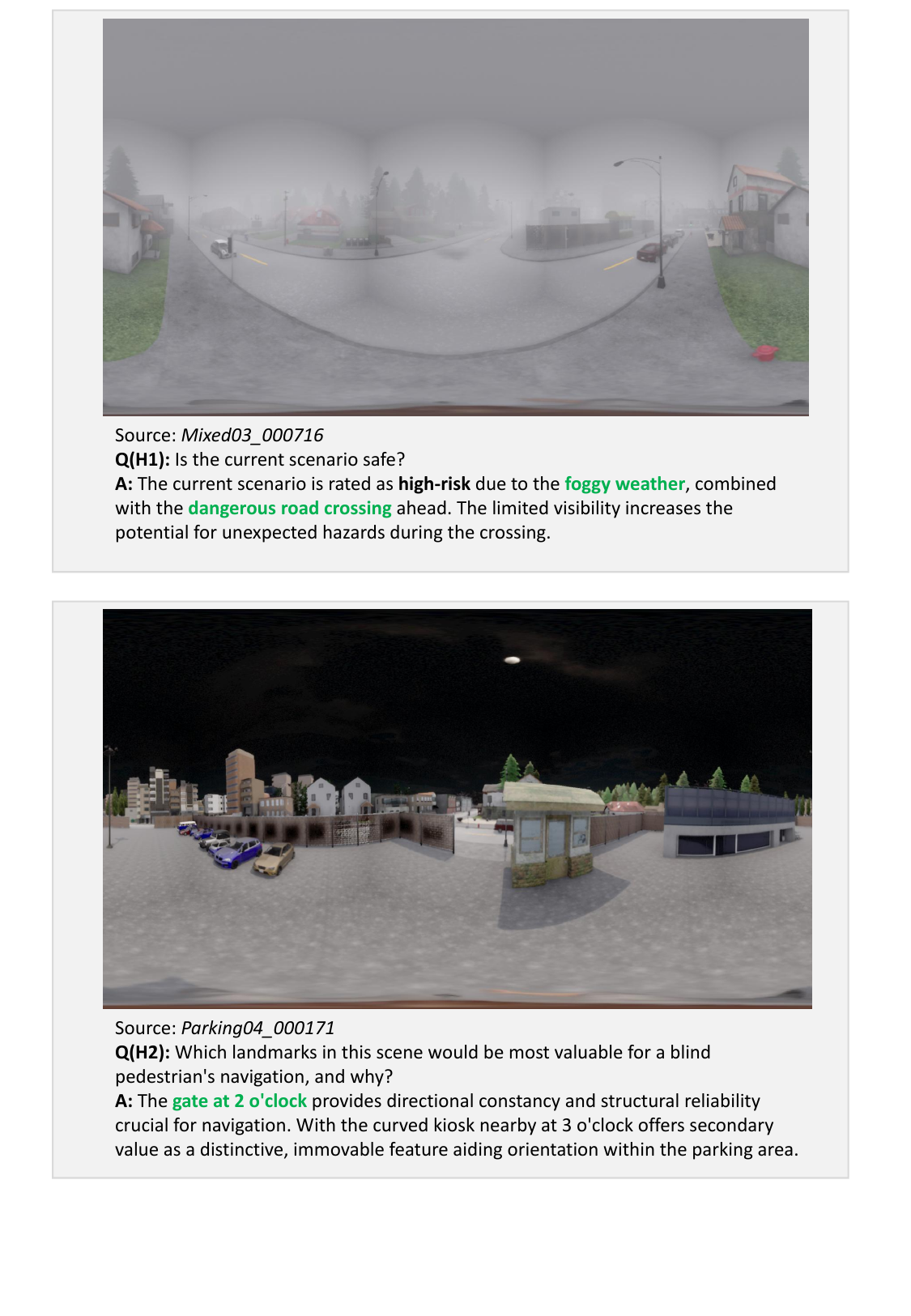}
    \caption{Examples of hard level VQA pairs}
    \label{fig:examplehard}
\end{figure}

\newpage

\begin{table}[H]
\centering
\caption{Resource List}
\label{resource}
\begin{tabular}{@{}ll@{}}
\toprule
\textbf{Resource} & \textbf{Content} \\ \midrule
CPUs & Intel Xeon Platinum 8368 \\
CPU Sockets per node & 2 \\
CPU Cores per node & 76 \\
CPU Threads per node & 152 \\
Cache L1 & 64K (per core) \\
Cache L2 & 1MB (per core) \\
Cache L3 & 57MB (shared, per CPU) \\
Main memory & 512 GB \\
Accelerators & 4x NVIDIA A100-40 \\
Memory per accelerator & 40 GB \\
Local disks & 960 GB NVMe SSD \\
Interconnect & InfiniBand HDR \\ \bottomrule
\end{tabular}%
\end{table}

\begin{figure}[H]
\centering
\begin{tcolorbox}[
  width=\textwidth,
  colback=gray!5,
  colframe=black!70,
  arc=2mm,
  boxrule=0.4pt,
  left=1pt, right=1pt, top=1pt, bottom=1pt
]
\begin{lstlisting}[
  basicstyle=\rmfamily,
  breaklines=true,
  breakindent=0pt,
  columns=fullflexible
]
internvl/train/internvl_chat_finetune.py 
  --model_name_or_path "pretrained/InternVL2-8B" 
  --conv_style "internlm2-chat" 
  --force_image_size 448
  --max_dynamic_patch 6
  --down_sample_ratio 0.5 
  --freeze_llm True 
  --freeze_mlp True 
  --freeze_backbone True 
  --use_llm_lora 16 
  --vision_select_layer -1 
  --dataloader_num_workers 4 
  --bf16 True 
  --num_train_epochs 4 
  --per_device_train_batch_size 
  --gradient_accumulation_steps 
  --save_strategy "steps" 
  --save_steps 200 
  --save_total_limit 1 
  --learning_rate 4e-5 
  --weight_decay 0.05 
  --warmup_ratio 0.03 
  --lr_scheduler_type "cosine" 
  --logging_steps 1 
  --max_seq_length 4096 
  --do_train True 
  --grad_checkpoint True 
  --group_by_length True 
  --dynamic_image_size True 
  --use_thumbnail True 
  --ps_version 'v2' 
  --deepspeed "zero_stage1_config.json" 
  --report_to "tensorboard" 
\end{lstlisting}
\end{tcolorbox}
\caption{Finetune Parameter}
\label{fig:finetune}
\end{figure}

\newpage
\begin{figure}[H]
\centering
\begin{tcolorbox}[
  width=\textwidth,
  colback=gray!5,
  colframe=black!70,
  arc=2mm,
  boxrule=0.4pt,
  left=1pt, right=1pt, top=1pt, bottom=1pt
]
\begin{lstlisting}[
  basicstyle=\rmfamily,
  breaklines=true,
  breakindent=0pt,
  columns=fullflexible
]
messages = [
        {
            "role": "system",
            "content":
                "You are an intelligent evaluator designed to evaluate the correctness and similarity of generative outputs for question-answer pairs. "
                "Your task is to compare the model prediction answer with the correct answer and determine if they match in meaning. Here's the scoring criteria:\n\n"
                "### Scoring Criteria:\n"
                "5 = Perfect match or Correct in meaning\n"
                "4 = Key information correct, minor flaws\n"
                "3 = Partially correct\n"
                "2 = Mostly wrong answer for key query, but some relevance\n"
                "1 = Completely wrong or nonsense sentences\n\n"
                "Your response must ONLY be the integer score (e.g., 4). DO NOT include any text or explanation."
        },
        {
            "role": "user",
            "content":
                f"Question: {question}\n"
                f"Correct Answer: {gt_answer}\n"
                f"Predicted Answer: {pred_answer}\n\n"
                "Please provide a score from 1 to 5 based on how well the predicted answer matches the correct answer."
        }
    ]
\end{lstlisting}
\end{tcolorbox}
\caption{Evaluation Prompt}
\label{fig:eval_gpt}
\end{figure}

\end{document}